\pdfoutput=1
\documentclass[11pt]{article}

\usepackage[utf8]{inputenc}
\usepackage[T1]{fontenc}
\usepackage{lmodern}
\usepackage{amsmath,amssymb}
\usepackage{graphicx}
\usepackage{booktabs}
\usepackage{array}
\usepackage{siunitx}
\usepackage{caption}
\usepackage{subcaption}
\usepackage[margin=1in]{geometry}
\usepackage{xcolor}
\usepackage{microtype}
\usepackage[colorlinks=true,linkcolor=blue,citecolor=blue,urlcolor=blue]{hyperref}
\usepackage{orcidlink}

\graphicspath{{figures/}}

\title{\textbf{A Fixed-Radius Distance-Band Benchmark for\\
Dimensionality-Reduction Fidelity}}

\author{
  Yoshio Takaeda\,\orcidlink{0009-0005-4968-9456}\\
  \textit{toor Inc.}\\
  \texttt{takaeda@toor.jpn.com}
}

\date{\today}

\begin{document}
\maketitle

\begin{abstract}
Dimensionality-reduction (DR) methods are routinely judged by how well each
point's $k$ nearest neighbors survive the 2-D embedding
(recall@$k$, trustworthiness, continuity). We argue this family is a
\emph{biased} measure of distance fidelity: it uses a per-point
\emph{variable radius} and a hard inclusion threshold, both of which
structurally favor neighbor-graph methods (t-SNE, UMAP) and penalize
methods that preserve absolute distances. We instead score DR fidelity with
a \emph{fixed-radius distance-band Shepard $\rho$}: the Spearman correlation
between high-D and 2-D pairwise distances, restricted to cumulative distance
bands so that near (within-group) and global (overall-layout) structure are
reported separately, with every point judged on the same absolute radius.
On synthetic datasets with known ground-truth geometry
(non-uniform density, dense clusters, a continuous closed-loop transition,
off-subspace outliers, and imbalanced two-population data),
at a realistic noise level (SNR${=}1$, $D{=}768$, $N{=}1000$), we benchmark
eight methods --- PCA, Isomap, t-SNE, UMAP, PyMDE, PCC, DREAMS, and the
closed-source \textbf{toorPIA} --- and show that (i) high global Shepard
$\rho$ can coexist
with a ${\approx}93\times$ collapse of within-cluster scale, invisible to
rank-based scores but obvious in a value-based over-compression metric;
(ii) recall@$k$ and the fixed-radius band disagree systematically, in the
direction the bias predicts; (iii) a membership-restricted Shepard $\rho$
resolves single-point and minority-population questions that many-pair
statistics cannot --- questions on which even DREAMS, the recent
local-plus-global hybrid that takes the near band on every dataset here,
fails silently. A supplementary out-of-sample (\emph{addplot}) test poses
the operational monitoring question --- does a never-seen anomaly land outside
the normal region, and does its direction identify its source? All metrics are
computed exactly on all pairwise distances, independently of any method's
internals; every number is reproducible offline, including for the
closed-source method, whose \emph{output coordinates} (not its algorithm) are
committed to the artifact. The benchmark is released as a
reproducible, externally citable characterization.
\end{abstract}

\section{Introduction}
\label{sec:intro}

Dimensionality reduction (DR) is how practitioners \emph{look at}
high-dimensional data: each feature vector becomes a dot in a 2-D map, and
the analyst reads groups, transitions, outliers, and minority populations off
the picture \cite{vandermaaten2008tsne,mcinnes2018umap,espadoto2019survey}.
Information is inevitably lost in the projection, so the value of the picture
rests on one question: how faithfully do the 2-D distances between points
reproduce the original high-dimensional distances --- separately for
\emph{near} pairs (fine, within-group structure) and \emph{far} pairs (the
overall layout)? On real data the true structure is unknown and maps end up
judged by eye; on synthetic data with a known generating geometry, every
method can be scored against ground truth. That is what this benchmark does.

The classic quantitative readout is the Shepard diagram \cite{shepard1962}
and its scalar summary, the Shepard $\rho$: the rank correlation between
high-D and 2-D distances over all point pairs. The catch is \emph{distance
concentration} \cite{beyer1999nearest,aggarwal2001surprising}: in high
dimensions almost every pair sits at a similar mid-to-far distance, and only
a thin sliver of pairs is genuinely close. A global $\rho$ computed over all
pairs is therefore dominated by far pairs, and the accuracy of near
distances --- the very structure the analyst zooms into --- is buried. A
method can crush every cluster to a featureless blob and still post a
near-perfect global $\rho$.

The field's standard remedy is to add a \emph{local} metric: recall@$k$,
trustworthiness, continuity \cite{venna2006trustworthiness,
lee2009qualityassessment} --- for each point, whether its $k$ nearest
neighbors survive the embedding. We argue in \S\ref{sec:metrics} that this
family is a \emph{biased} measure of near-distance fidelity: it judges every
point on its own $k$-NN radius (a per-point \emph{variable} radius) with a
hard 0/1 inclusion threshold, and both choices structurally favor the
neighbor-graph methods that optimize $k$-NN objects directly, while
penalizing methods that preserve absolute distances. The result is an
uncomfortable status quo: global structure is judged by a metric blind to
near structure, and near structure by a metric biased toward one family of
methods.

This paper scores DR fidelity instead with a \emph{fixed-radius
distance-band} Shepard $\rho$ --- near (the first mode of the pairwise
distance profile) and global (all pairs) reported separately, every point
judged on the same absolute radius --- complemented by value-based metrics
(band stress, a tightest-cluster over-compression factor) that catch what
rank statistics are structurally blind to, and by membership-restricted
variants of the same statistic for single-point (outlier) and
minority-population questions. On five synthetic datasets with known
geometry at a realistic noise level (SNR${=}1$, $D{=}768$, $N{=}1000$), we
benchmark eight methods, including the closed-source toorPIA, and release
every number as an offline-reproducible, externally citable
characterization --- including the closed-source method's, whose output
coordinates (not its algorithm) are committed to the artifact.

\paragraph{Contributions.}
\begin{enumerate}
  \item \textbf{A fixed-radius distance-band fidelity metric.} We formalize the
  band-restricted Shepard $\rho$ (near band = first mode of the distance
  profile; global = all pairs) as a \emph{fair}, fixed-radius alternative to
  recall@$k$/trustworthiness/continuity, and give the structural argument
  (variable radius $+$ hard threshold) for why the latter are biased toward
  neighbor-graph methods (\S\ref{sec:metrics}).
  \item \textbf{An eight-method benchmark on known-geometry synthetic data} at a
  realistic noise level, separating rank-based fidelity (Shepard $\rho$) from
  value-based fidelity (stress, within-cluster over-compression), and exposing
  a scale-collapse failure mode that global $\rho$ hides (\S\ref{sec:results}).
  \item \textbf{Membership-restricted scoring} for single-point (outlier) and
  minority-population questions, plus an operational out-of-sample
  (\emph{addplot}) monitoring criterion (\S\ref{sec:outliers},
  \S\ref{sec:populations}, \S\ref{sec:addplot}).
  \item \textbf{A reproducible, externally citable characterization of a
  closed-source method} (toorPIA), designed so that its published claims do not
  rest on the vendor's judgment (\S\ref{sec:reproducibility}).
\end{enumerate}

\section{Related Work}
\label{sec:related}

\subsection{Dimensionality-reduction methods}

The methods compared here span the main families in use. PCA
\cite{hotelling1933pca} is the linear baseline: an orthogonal projection
maximizing retained variance. Isomap \cite{tenenbaum2000isomap} performs
classical scaling on graph-geodesic distances; locally linear embedding
\cite{roweis2000lle} and Laplacian eigenmaps \cite{belkin2003laplacian}
embed by solving spectral problems on neighborhood graphs. t-SNE
\cite{vandermaaten2008tsne} and UMAP \cite{mcinnes2018umap} --- today's
default visualization choices --- optimize agreement between neighbor
distributions or graphs built in the two spaces, which preserves local
neighborhoods while leaving large-scale distances only weakly constrained.
PyMDE \cite{agrawal2021pymde} is a general minimum-distortion embedding
framework, configured here with an absolute-distance loss (a modern
stress-based MDS). PCC (``Preserving Clusters and Correlations'')
\cite{pcc2025} pairs a cluster-observability term with a correlation
objective between high-D and 2-D distances to a sampled reference set ---
the latter a distance-fidelity objective with sparse constraints
(\S\ref{sec:methodsconfig} states the configuration used here). DREAMS
\cite{kury2026dreams} addresses the local--global trade-off head on: it
augments t-SNE's neighbor objective with a coordinate-level regularization
pull toward the PCA embedding, weighted by a single strength parameter
$\lambda$, and reports strong preservation of both local and global
structure --- precisely the two readings this benchmark separates, which
makes it a natural inclusion. toorPIA is a closed-source commercial
method described by its vendor as related to the spectral
Laplacian-eigenmaps family \cite{belkin2003laplacian}; its internal
algorithm is not public and is not verified here --- this paper
characterizes its input--output behavior only
(\S\ref{sec:reproducibility}).

\subsection{Evaluating embedding fidelity}

The oldest fidelity readouts are value-based and global: the Shepard diagram
\cite{shepard1962} plots 2-D against high-D distances for all pairs, and
Kruskal's stress \cite{kruskal1964mds} summarizes the value error. The
modern DR literature instead standardized on \emph{local, rank-based}
metrics --- trustworthiness and continuity
\cite{venna2006trustworthiness}, unified with recall-type scores in the
co-ranking framework of \cite{lee2009qualityassessment} --- and large-scale
quantitative surveys score methods on precisely this family
\cite{espadoto2019survey}. Cluster-level metrics extend the family upward:
Steadiness and Cohesiveness \cite{jeon2021steadiness} score how faithfully
\emph{inter-cluster} structure survives the projection, though their cluster
extraction is itself built on neighbor graphs, placing them in the same
methodological family whose bias we analyze below. That metric
\emph{selection} itself skews DR evaluation is by now an explicit concern:
\cite{bae2025metric} show that commonly co-selected metrics are empirically
correlated in ways their design intent does not predict, biasing evaluations
toward particular method families, and propose selecting metrics by measured
behavior instead --- the same concern, at the level of metric suites, that
\S\ref{sec:recallbias} raises about the internals of the $k$-NN family. Our
band-restricted Shepard $\rho$ stays in the
rank-based tradition, but it partitions pairs by \emph{absolute distance} on
the global pair-distance distribution rather than by each point's neighbor
ranks: the near band is one fixed radius applied to every point, and near
and global fidelity are reported as separate numbers. The $k$-NN family is
retained as a labelled reference, and its variable-radius, hard-threshold
bias is analyzed in \S\ref{sec:recallbias}.

\subsection{Distance concentration in high dimensions}

That high-dimensional distances concentrate --- the contrast between the
nearest and farthest neighbor shrinks as dimension grows, and the
meaningfulness of ``nearest'' itself degrades --- is classical
\cite{beyer1999nearest,aggarwal2001surprising}. Its consequence for
\emph{evaluation}, which this paper acts on, has drawn less attention: when
most pairs of a high-D dataset sit in a narrow far band, any all-pair
statistic is dominated by far pairs, so a global Shepard $\rho$ is
structurally blind to near-distance errors. The fixed-radius near band of
\S\ref{sec:bands} is the direct remedy. (Our datasets are
redundancy-rich by design, so concentration shapes the pair-distance
profile without destroying the signal; \S\ref{sec:protocol}.)

\section{Metrics: fixed-radius bands vs.\ variable-radius $k$-NN}
\label{sec:metrics}

\subsection{Shepard $\rho$ and why the global number hides near structure}
\label{sec:bands}

The classic distance-fidelity readout is the Shepard diagram
\cite{shepard1962} and its scalar summary, the \emph{Shepard $\rho$}: the
Spearman rank correlation between the high-D and 2-D distances of all
$N(N{-}1)/2$ point pairs. Throughout this benchmark, every metric is computed
exactly on all pairwise distances, independently of any method's internal
approximations (reference-point sampling, neighbor graphs); high-D distance
is the dataset's defined (Euclidean) distance, and 2-D distance is Euclidean.

\begin{figure}[t]
  \centering
  \includegraphics[width=0.8\linewidth]{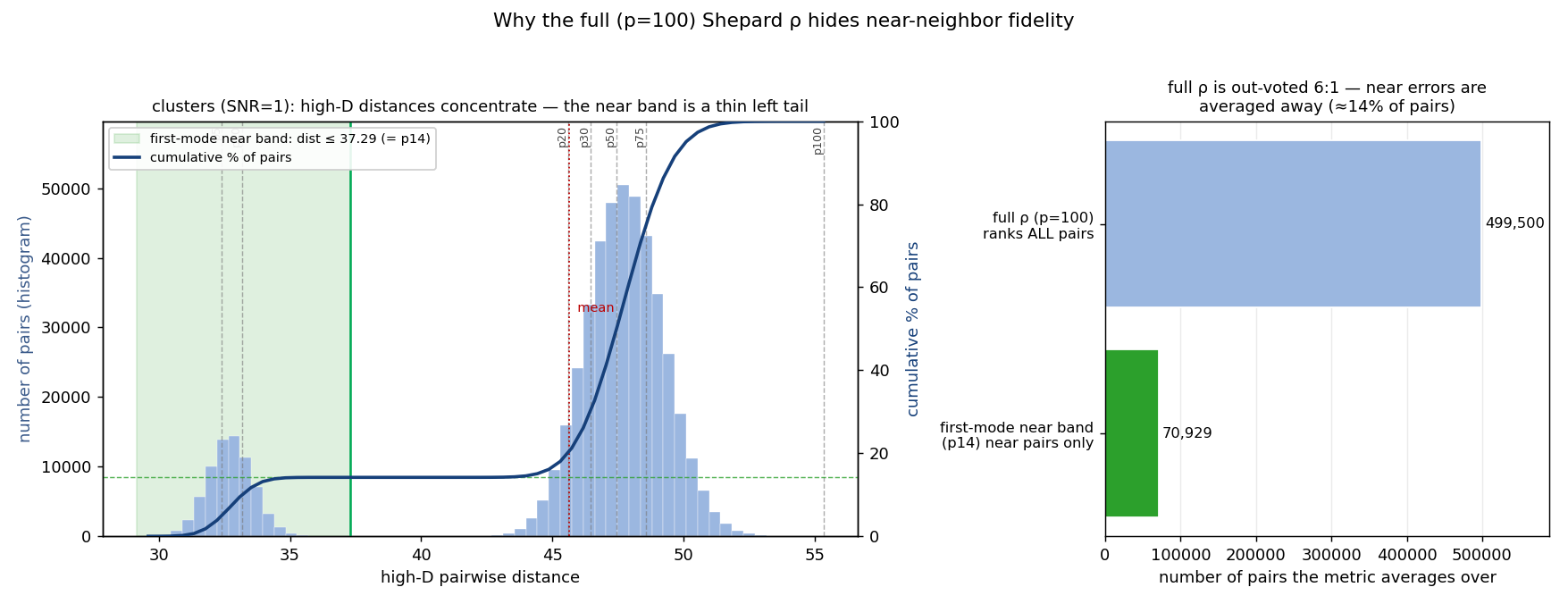}
  \caption{High-D pairwise-distance profile (clusters dataset). Most pairs sit
  in the far mode (between-group); the near band (first mode, within-group
  fine structure) is the small left tail that a global Shepard $\rho$
  averages away.}
  \label{fig:distdist}
\end{figure}

The catch is distance concentration \cite{beyer1999nearest,
aggarwal2001surprising}: in high dimensions almost every pair of points sits
at a similar mid-to-far distance, and only a thin sliver of pairs is
genuinely close. On the clusters dataset (Fig.~\ref{fig:distdist}) the
within-cluster pairs are ${\approx}14\%$ of the roughly $5{\times}10^5$
pairs; the rest pile up in the far mode. Because the full $\rho$ ranks all
pairs together, near-distance errors are out-voted roughly $6{:}1$ and
averaged away --- a method can crush every cluster to a featureless blob and
still post a near-perfect global number. We therefore report the full $\rho$
as what it is --- a global-structure metric --- and score near-distance
fidelity separately.

We do so with cumulative \emph{distance bands} defined on the global high-D
pairwise-distance distribution: the band at cutoff $p$ holds the pairs whose
high-D distance lies in the lowest $p\%$ of all pairs, and the band-restricted
Shepard $\rho$ is the same Spearman statistic computed within that subset.
Sweeping $p \in \{5,10,20,30,50,75,100\}$ traces a near$\to$far profile
(committed in the artifact), and $p{=}100$ recovers the classic global
number. The headline \emph{near band} is structure-adaptive rather than a
fixed percentile. The pairwise-distance profile of structured data is
multimodal --- its first mode is the within-structure pairs --- and the near
band is all pairs up to the density valley where that first mode decays into
the tail. The estimator (\texttt{metrics.distances.first\_mode\_threshold};
deterministic, computed on all pairs, constants disclosed) histograms the
distances into 256 equal-width bins, smooths twice with a length-9 boxcar,
and takes the profile minimum between the first two local maxima, where a
mode must reach at least 5\% of the profile maximum, the valley must dip
below 95\% of the first mode (so tail-noise bumps do not count as modes), and
the boundary must lie below the median distance; if no second mode exists,
the band falls back to the 5th-percentile radius (flagged in the committed
CSVs; never triggered on these datasets). At SNR${=}1$ the detected boundary
lands at p20.5 / p14.2 / p14.9 / p19.7 / p18.0 on density / clusters /
transition / outliers / populations, and on the three clustered datasets
(clusters, outliers, populations) it coincides with the true within-cluster
pair fraction ($14.2\%$ / $19.7\%$ / $18.0\%$): the label-free band recovers
the ground-truth notion of ``near''.

Crucially, the band is one absolute distance threshold for the whole
dataset --- every point is judged on the same radius. This fixed-radius
property is what makes the near band a fair near-neighbor metric, and it is
exactly the property the field's standard local metrics lack.

\subsection{Why recall@$k$ / trustworthiness / continuity are a biased reference}
\label{sec:recallbias}

\begin{figure}[t]
  \centering
  \includegraphics[width=0.9\linewidth]{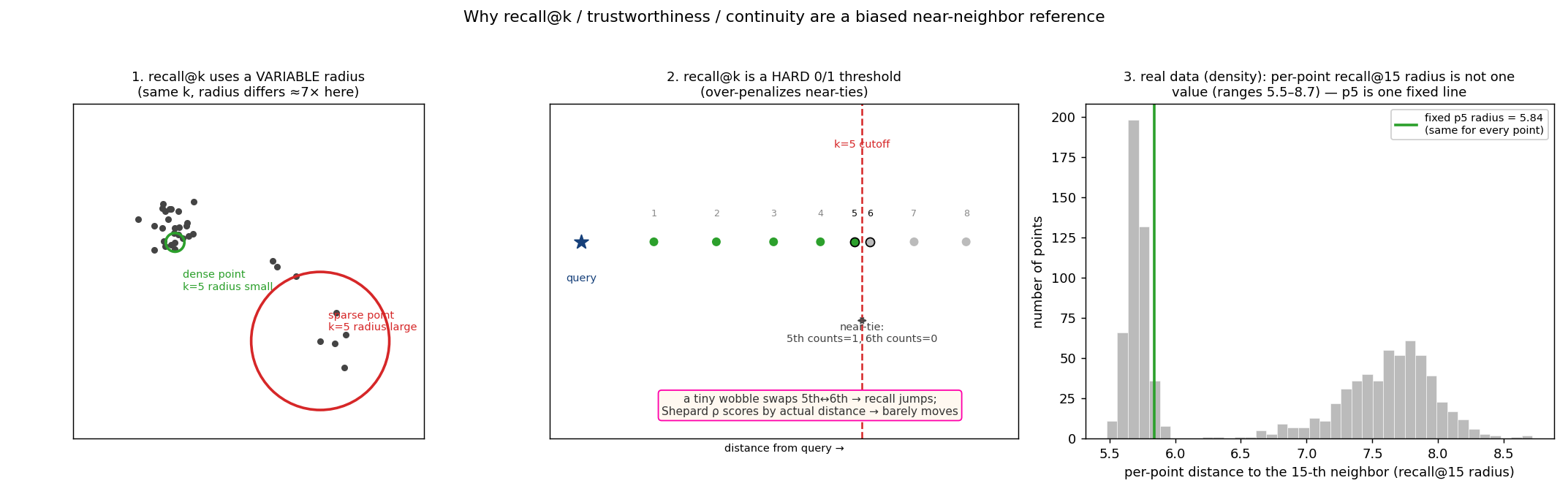}
  \caption{Why recall@$k$ is a biased reference (density dataset). It judges
  each point on its own $k$-NN radius (panels 1, 3) with a hard in/out cutoff
  (panel 2); the fixed-radius band (green) treats all points identically.}
  \label{fig:recallbias}
\end{figure}

The DR literature's standard local metrics are recall@$k$ --- for each point,
the fraction of its $k$ high-D nearest neighbors that remain among its $k$
nearest in 2-D --- and its cousins trustworthiness and continuity
\cite{venna2006trustworthiness,lee2009qualityassessment}. Two structural
choices make this family a biased test of near-\emph{distance} fidelity
(Fig.~\ref{fig:recallbias}). (1)~\emph{Variable radius}: with $k$ fixed, a
point in a dense region encloses its $k$ neighbors within a tiny radius,
while a point in a sparse region needs a much larger one, so every point is
judged on a different distance scale. (2)~\emph{Hard 0/1 threshold}: the
$k$-th and $(k{+}1)$-th neighbors can be nearly equidistant, yet one counts
fully and the other not at all; a tiny coordinate wobble flips membership and
the score jumps, although the actual distances barely moved. Both choices are
structurally favorable to methods that optimize a $k$-NN neighborhood object
directly (t-SNE, UMAP \cite{vandermaaten2008tsne,mcinnes2018umap}): the
metric then measures agreement with a $k$-NN construction, not faithful
reproduction of near distances. The fixed-radius near band of
\S\ref{sec:bands} has neither problem --- one radius for all points, scored
by a continuous rank correlation on the actual distances.

We do not discard these metrics. Recall@$k$ ($k \in \{5,15,30\}$),
trustworthiness, and continuity are computed and reported throughout as a
clearly labelled \emph{reference} block --- kept for comparability with the
literature, deliberately excluded from the composite ranking
(\S\ref{sec:composite}). Empirically, the two blocks disagree in exactly the
direction the bias predicts: distance-preserving methods dominate the
fixed-radius bands, while t-SNE and UMAP dominate recall@$k$
(\S\ref{sec:results}).

Two caveats, stated openly. A per-point band-Shepard variant (each point's
own lowest-$p\%$ pairs) is committed in the artifact as a secondary view, but
it re-introduces the variable radius, so the fixed-radius global band remains
primary. And on non-uniform-density data the global first-mode band holds
more pairs where points are dense, so the near $\rho$ is weighted toward
near-structure in dense regions; the trade-off is point-uniformity versus
fixed-radius fairness, and we choose fairness.

\subsection{Complementary value-based and membership-restricted metrics}
\label{sec:valuemetrics}

Rank correlation is scale-invariant: it cannot see a method that preserves
distance \emph{order} while distorting distance \emph{values}. Two
value-based complements close this gap, and a membership-restricted variant
of the same Shepard machinery covers questions that no distance-percentile
band can ask.

\paragraph{Band stress.} Within each band we report the normalized stress
\cite{kruskal1964mds}
\[
  \mathrm{stress} \;=\; \sqrt{\frac{\sum_{ij} (a_{ij} - \alpha\, b_{ij})^2}
                                   {\sum_{ij} a_{ij}^2}},
  \qquad
  \alpha = \frac{\langle a, b\rangle}{\langle b, b\rangle},
\]
where $a$ are the high-D and $b$ the 2-D distances of the band's pairs and
$\alpha$ is the optimal global scale. Fitting $\alpha$ first matters:
\cite{smelser2024stress} show that plain normalized stress is sensitive to a
uniform rescaling of the projection --- an operation that changes nothing
about the picture --- and propose evaluating stress at the optimal scaling as
the correction; the least-squares $\alpha$ above is exactly that
scale-invariant formulation, applied per band. Stress catches methods that
keep the
distance ordering but distort the values (e.g., scale-free objectives);
$\mathrm{fidelity} = \max(0,\, 1 - \mathrm{stress})$ is also committed.

\paragraph{Tightest-cluster scale (over-compression $\times$).} The structure
that a global-layout trade-off sacrifices first is the tightest cluster. Once
per dataset, in truth space and hence identically for every method, we select
the labelled cluster (of at least 20 points) with the smallest median
within-cluster pairwise distance, and compare its scale relative to the
overall spread in each space:
\[
  \mathrm{ratio}(\text{space}) =
  \frac{\operatorname{median}(\text{within-cluster pair distances})}
       {\operatorname{median}(\text{all pair distances})},
  \qquad
  \text{over-compression} \times =
  \frac{\mathrm{ratio}(\text{truth})}{\mathrm{ratio}(\text{2-D})}.
\]
A value ${\approx}1$ means the tightest cluster keeps its relative scale;
${\gg}1$ means it is crushed toward a point; ${\ll}1$ means it is inflated.
Crushing is the real harm --- a crushed cluster's internal structure cannot
be read back from the map, whereas inflation is legible, if exaggerated ---
so the failure flag marks only the crush side (the table's worst case when it
exceeds $5\times$).

\paragraph{Membership-restricted $\rho$.} The outliers and populations
datasets ask questions about designated \emph{subsets} of points. We answer
them with the same standard statistic, restricting the pair set by
\emph{endpoint membership} instead of by distance percentile: the
anomaly-pair $\rho$ (``outlier $\rho$'': all pairs with at least one
ground-truth outlier endpoint), and the population diagnostics
(majority-internal, minority-internal, and cross-population $\rho$;
\S\ref{sec:populations}). The restriction exists because such violations
involve few pairs --- a single outlier participates in $O(1/N)$ of all
pairs --- so the all-pair $\rho$ barely moves even under complete failure.

\paragraph{Non-circularity.} No included method is optimized on a quantity we
score it with. In particular, PCC is run with its published Pearson
(value-based) loss \cite{pcc2025} while the primary metric is the Spearman
(rank-based) Shepard $\rho$: optimizing Spearman would be teaching to the
test, whereas optimizing Pearson and scoring well on the rank metric is the
honest, non-circular outcome. The other objectives are likewise disjoint from
the scored metrics (t-SNE/UMAP optimize neighbor embeddings, PCA variance,
Isomap geodesic MDS); PyMDE's absolute-distance value loss is the closest in
spirit to the \emph{secondary} stress metric, which we flag explicitly.

\subsection{Composite ranking}
\label{sec:composite}

Each dataset's ranking table scores \emph{distance fidelity only}. The
methods are ordered twice --- by full $\rho$ and by first-mode near $\rho$
(both vs.-ambient) --- and each order awards 5 points to the 1st place down
to 1 point to the 5th (6th--8th receive 0). The composite $\Sigma$ is the
sum of the two, and rows are sorted by $\Sigma$. The outliers and
populations datasets each add a third scored column on a
membership-restricted $\rho$ (same $5\ldots1$ scheme, $\Sigma$ = full +
near + third): the anomaly-pair $\rho$ on outliers and the minority-pair
$\rho$ (pairs with at least one minority endpoint) on populations, so a
method cannot rank well there while failing the single-point or the
minority question. The $k$-NN reference block is deliberately
unscored (\S\ref{sec:recallbias}).

Aggregation follows one convention throughout: stochastic methods run
$R{=}3$ seeds and report the median with a bootstrap 95\% CI; PCA, Isomap,
DREAMS, and toorPIA are deterministic and show point values. When two
methods' CIs overlap on a metric, no strict winner is asserted. Outright failures are
flagged in the tables independently of rank: a negative near-band $\rho$, the
table's worst tightest-cluster crush when it exceeds $5\times$, and a
negative anomaly-pair $\rho$.

\section{Datasets and Experimental Setup}
\label{sec:datasets}

\subsection{Generation protocol and ground truth}
\label{sec:protocol}

Every dataset is built in a low-dimensional \emph{latent} space, where its
geometry is unambiguous, and mapped into the ambient dimension $D$ by a
random \emph{orthonormal} projection. An orthonormal map is an isometry, so
the ground-truth distance equals the clean ambient Euclidean distance.
SNR-controlled isotropic Gaussian noise is then added in all $D$ dimensions
to form the features $X$ that the methods embed. Because truth and ambient
differ only by that noise, every metric is computed both \emph{vs.-truth}
(against the clean generating distances) and \emph{vs.-ambient} (against the
noisy distances the method actually saw). The canonical configuration is
$D{=}768$, $N{=}1000$, SNR${=}1$, with $R{=}3$ seeds for the stochastic
methods; runs are CPU-bound and single-threaded, every stochastic step is
seeded, and re-running with the same arguments reproduces identical numbers.
The report fixes a single noise level, SNR${=}1$, because it is the most
discriminative: at SNR${=}\infty$ even simple linear methods look good, while
realistic noise separates the methods.

This design is deliberately \emph{redundancy-rich}: the orthonormal
projection spreads each latent coordinate over all $D$ ambient columns, so
the ambient data is in effect 768 noisy re-measurements of ${\sim}10$ latent
quantities. In every pairwise distance the signal contributions add
coherently while the isotropic noise self-averages, so the ambient dimension
is nominal and no curse of dimensionality operates --- by construction. This
is the noise-friendly extreme. The artifact additionally contains a
noise-dims dimension sweep probing the opposite, redundancy-free regime
(signal confined to 3 columns, every further column pure noise, effective
SNR $= 3/(D{-}3)$); rankings need not transfer between the two regimes, and
we refer the reader to the repository for that supplement.

\subsection{Methods and configuration}
\label{sec:methodsconfig}

The eight methods are PCA \cite{hotelling1933pca}, Isomap
\cite{tenenbaum2000isomap}, t-SNE \cite{vandermaaten2008tsne}, UMAP
\cite{mcinnes2018umap}, PyMDE \cite{agrawal2021pymde}, PCC \cite{pcc2025},
DREAMS \cite{kury2026dreams},
and the closed-source toorPIA, all producing 2-D embeddings of the same
input $X$. The open-source methods run at library defaults (t-SNE
perplexity 30, UMAP \texttt{n\_neighbors} 15; PyMDE with an
absolute-distance loss). PCC runs \emph{label-free}
(\texttt{cluster=False}): as shipped, its cluster-observability term
consumes caller-supplied cluster assignments, which no method receives in
this benchmark, so PCC is scored on its published Pearson correlation
objective alone, with the reference set at its maximum size ($N$, sampled
with replacement). Every claim about PCC in this paper refers to this
label-free configuration. DREAMS runs at its authors' published defaults:
the 2-component PCA embedding serves as both the initialization and the
regularization target, with strength $\lambda{=}0.15$, perplexity 30, and
Barnes--Hut gradients, via the authors' openTSNE fork (the fork's current
head crashes every plain \texttt{fit()} call --- an unused parameter leaks
into its optimizer --- so the benchmark wrapper drops the dead parameter at
import time and the installed package remains the published source,
documented in \texttt{methods/dreams\_method.py}). With the fixed PCA
initialization and single-threaded gradient descent DREAMS has no remaining
randomness --- three seeds produce byte-identical embeddings --- so it is
treated as deterministic. toorPIA is called through its embedding endpoint
(\texttt{basemap\_embedding}, per-row L2 normalization disabled), so it
embeds the very same raw vectors the other methods see; the endpoint exposes
no random seed and is deterministic, and its output coordinates are cached
and committed for offline reproduction (\S\ref{sec:reproducibility}). A
committed hyperparameter-sensitivity sweep (t-SNE/UMAP's neighborhood size
over 5--100; DREAMS's $\lambda$ over 0.05--0.5) shows that no setting
changes a composite leader outright, with two genuine sensitivities
disclosed in the artifact: tuned t-SNE (perplexity 100) overtakes the
near-band $\rho$ leaders on density and draws level with toorPIA's density
composite, at the cost of its own recall@15; and DREAMS's $\lambda$ acts as
a real local--global dial on density (full $\rho$ 0.39 $\to$ 0.62 as
$\lambda$ rises to 0.5, still below toorPIA's 0.82, with its near-band lead
kept at every $\lambda$).

\subsection{The five datasets}
\label{sec:fivedatasets}

\paragraph{Non-uniform density.} A uniform region, a tight Gaussian core,
and a sparse spherical shell, with deliberately different densities
(effective dimension ${\approx}10$). It tests density distortion (do the
neighbor-graph methods inflate the dense core?), demonstrates the
recall@$k$ bias of \S\ref{sec:recallbias}, and probes near-band fidelity
where near pairs are concentrated in the dense core.

\paragraph{Distinct dense clusters.} $K{=}7$ small dense Gaussian clusters
placed on $K$ mutually orthogonal latent axes, so the cluster configuration
spans $K{-}1 = 6$ affine dimensions --- genuinely high-dimensional global
geometry that a 2-D linear projection cannot reproduce by construction. The
single knob is the dynamic range (inter-cluster distance $\div$
intra-cluster spacing). It tests whether a method preserves fine
within-cluster structure (near band) while also placing the clusters
correctly (global).

\paragraph{Continuous closed-loop transition.} $K{=}7$ dense typical-state
clusters at known centroids on mutually orthogonal axes, connected by a
continuous transition region parameterized by a cyclic $t$ that runs through
every centroid and closes into a loop ($0{\to}1{\to}\dots{\to}6{\to}0$). The
bridges are heterogeneous (their perpendicular spread widens toward the
middle), and the clusters are as dense as the density dataset's tight core.
Placing the centroids on orthogonal axes rather than on a circle is
deliberate: a circle makes the global geometry intrinsically 2-dimensional,
which a 2-component PCA reproduces trivially, whereas the orthogonal
configuration spans 6 affine dimensions (PCA's top two components capture
only ${\sim}38\%$ of the variance). It tests reproduction from near (dense
clusters) to far (the loop), and whether the transition stays continuous or
fragments.

\paragraph{Off-subspace outliers.} A bulk of $K{=}5$ dense clusters spanning
the first latent dimensions, plus 3 anomalous \emph{directions} $\times$ 2
near-duplicate outliers each: direction $j$ has its own dedicated latent
axis, orthogonal to the entire subspace the bulk spans, and its two members
sit on that axis at $3\,R_g$ and $3.1\,R_g$ ($R_g$ = the bulk's radius of
gyration; the $0.1\,R_g$ offset matches the bulk's own median
nearest-neighbor spacing). Off-subspace placement is the point: a sample
acquired under a different condition varies along feature directions the
bulk does not span, and a single off-subspace point carries only
$(3 R_g)^2/N$ of the variance, which a variance-truncating projection may
legally drop --- separating ``constrains distances'' from ``constrains
projected variance''. It tests whether a single far-away point keeps its
separation margin in 2-D, the single-point property that many-pair
statistics cannot resolve (\S\ref{sec:valuemetrics}). The separation factor
(default 3) is swept over $\{1.5, 2, 3, 5, 8\}$ in the artifact.

\paragraph{Imbalanced two populations.} A majority population (the
5-orthogonal-axes cluster recipe) and a minority with the \emph{same}
internal 5-cluster geometry, built in a disjoint latent block and offset so
that every cross-population cluster-center distance is exactly twice the
within-population one --- a strict two-level hierarchy. The canonical
setting is 95\% vs.\ 5\% (swept up to 50/50 in the artifact). The situation
is ubiquitous rather than adversarial: dominant production runs vs.\ a
rarely used operating mode, a large healthy cohort vs.\ a small patient
group with subtypes --- and the minority is very often the actual object of
the analysis, while its existence is unknown to the analyst in advance.
Extracting it from a map requires two readings positive at once: the
minority drawn as a recognizable separate group (cross-population $\rho$)
and a trustworthy internal structure (minority-internal $\rho$). A small
minority \emph{without} internal structure is the outliers dataset, so the
two datasets connect continuously.

\section{Results}
\label{sec:results}

All ranking tables report the committed v1.4.0 results at SNR${=}1$
($D{=}768$, $N{=}1000$); $\rho$ values are vs.-ambient (the noisy distances
the methods actually saw). The corresponding vs.-truth values are committed in
the artifact's per-run CSVs and lead to the same qualitative conclusions.

\subsection{Non-uniform density}
\label{sec:density}

\begin{table}[t]
\centering
\caption{Non-uniform density: distance-fidelity ranking at SNR${=}1$.\; Composite points: 1st$\to$5 \dots\ 5th$\to$1 on the full-$\rho$ order and on the near-$\rho$ order; rows sorted by $\Sigma$. $\rho$ columns are Shepard (Spearman) correlations vs.-ambient. \textbf{Bold} = best in column, \textit{italic} = worst; \textsuperscript{\dag} = outright-failure flag (negative near-band $\rho$, or worst tight-cluster crush exceeding $5\times$). Brackets are bootstrap 95\% CIs over $R{=}3$ seeds; deterministic methods (PCA, Isomap, DREAMS, toorPIA) show point values. recall/trust/continuity are the variable-radius $k$-NN reference block (biased; unscored). Values transcribed verbatim from the v1.4.0 committed results.}
\label{tab:density}
\resizebox{\textwidth}{!}{%
\begin{tabular}{lccccccccc}
\toprule
method & full & near & $\Sigma$ & full $\rho$ (global) & near $\rho$ (first-mode) & scale $\times$ & recall@15 & trust@15 & cont.@15 \\
\midrule
toorPIA & 4 & 4 & \textbf{8} & 0.815 & 0.312 & 0.438 & 0.067 & 0.627 & \textbf{0.821} \\
DREAMS & 1 & \textbf{5} & 6 & 0.585 & \textbf{0.363} & 0.239 & 0.245 & \textbf{0.776} & 0.804 \\
PCC & \textbf{5} & \textit{0} & 5 & \textbf{0.820 {\scriptsize[0.816, 0.824]}} & -0.028 {\scriptsize[-0.047, -0.002]}\,\textsuperscript{\dag} & 93.468 {\scriptsize[93.389, 93.583]}\,\textsuperscript{\dag} & 0.056 {\scriptsize[0.055, 0.059]} & 0.603 {\scriptsize[0.593, 0.604]} & 0.753 {\scriptsize[0.749, 0.759]} \\
PCA & 3 & 1 & 4 & 0.699 & 0.093 & \textbf{0.728} & 0.069 & 0.628 & 0.785 \\
PyMDE & 2 & 2 & 4 & 0.598 {\scriptsize[0.598, 0.608]} & 0.218 {\scriptsize[0.202, 0.254]} & 0.460 {\scriptsize[0.419, 0.516]} & \textit{0.043 {\scriptsize[0.041, 0.047]}} & \textit{0.568 {\scriptsize[0.567, 0.570]}} & \textit{0.634 {\scriptsize[0.632, 0.646]}} \\
t-SNE & \textit{0} & 3 & 3 & 0.300 {\scriptsize[0.277, 0.309]} & 0.260 {\scriptsize[0.259, 0.266]} & 0.215 {\scriptsize[0.214, 0.216]} & \textbf{0.254 {\scriptsize[0.250, 0.259]}} & 0.773 {\scriptsize[0.770, 0.779]} & 0.803 {\scriptsize[0.798, 0.803]} \\
Isomap & \textit{0} & \textit{0} & \textit{0} & 0.559 & -0.038\,\textsuperscript{\dag} & 0.383 & 0.060 & 0.601 & 0.776 \\
UMAP & \textit{0} & \textit{0} & \textit{0} & \textit{0.156 {\scriptsize[0.154, 0.162]}} & 0.055 {\scriptsize[0.040, 0.057]} & 0.279 {\scriptsize[0.275, 0.283]} & 0.210 {\scriptsize[0.206, 0.213]} & 0.733 {\scriptsize[0.729, 0.734]} & 0.813 {\scriptsize[0.812, 0.815]} \\
\bottomrule
\end{tabular}}%
\end{table}

\begin{figure}[t]
  \centering
  \includegraphics[width=\linewidth]{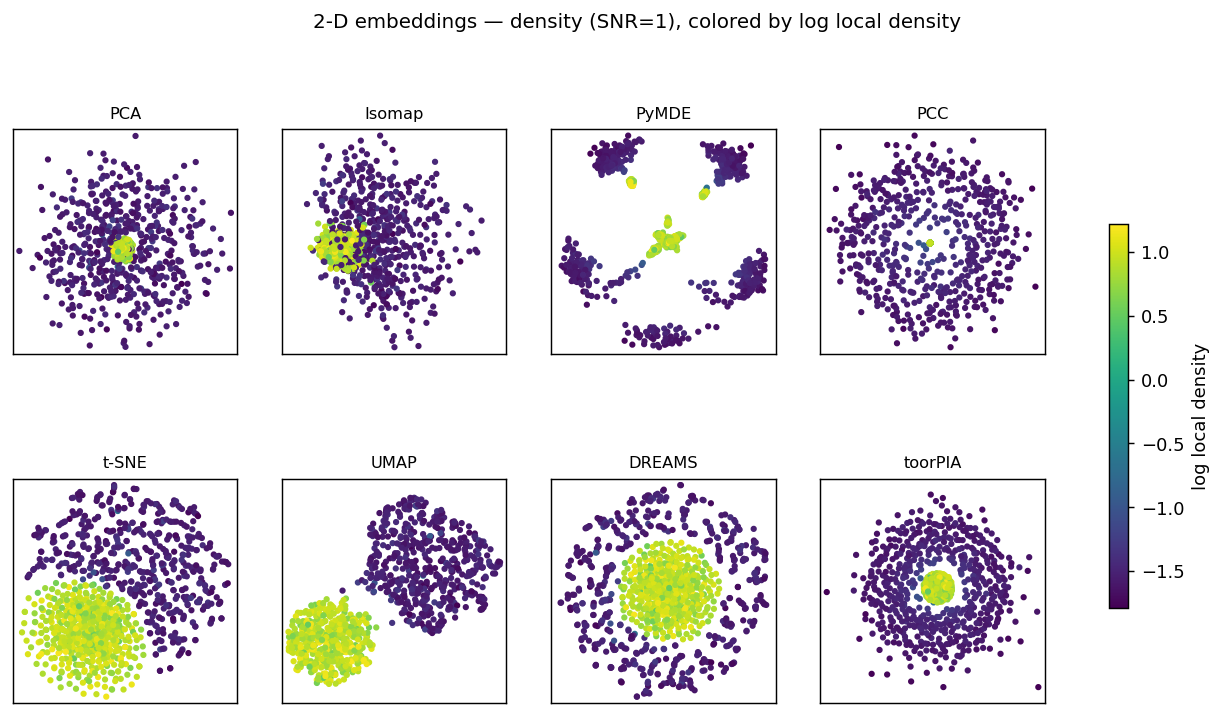}
  \caption{Density dataset embeddings (SNR${=}1$), colored by log local
  density. The neighbor-graph methods inflate the dense core into a
  dominant blob; PCC draws it as a near-point; toorPIA keeps a compact,
  legible core inside the sparse shell.}
  \label{fig:density_emb}
\end{figure}

\begin{figure}[t]
  \centering
  \begin{subfigure}{0.58\linewidth}
    \includegraphics[width=\linewidth]{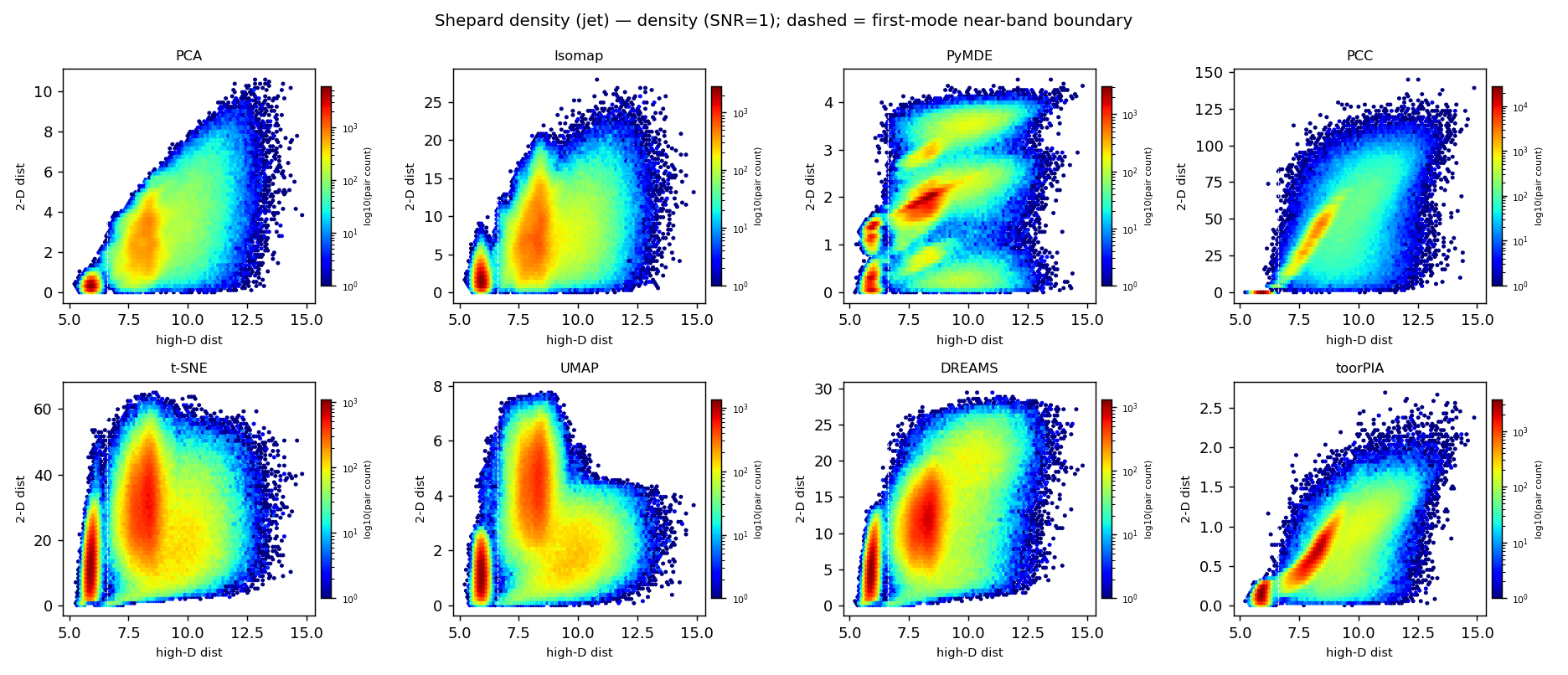}
    \caption{Shepard density (log pair count; dashed line $=$ first-mode
    near-band boundary).}
  \end{subfigure}\hfill
  \begin{subfigure}{0.40\linewidth}
    \includegraphics[width=\linewidth]{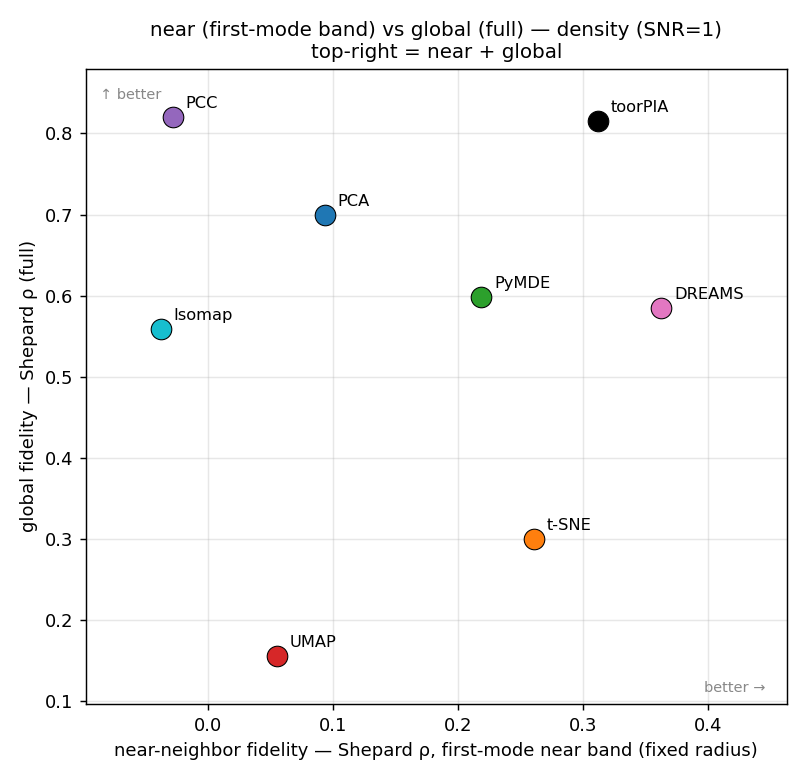}
    \caption{Near-band $\rho$ vs.\ global $\rho$; top-right preserves both.}
  \end{subfigure}
  \caption{Density dataset (SNR${=}1$): value-level and rank-level views of
  the same embeddings.}
  \label{fig:density_panels}
\end{figure}

The density dataset (uniform region $+$ tight Gaussian core $+$ sparse shell)
exposes the headline scale-collapse phenomenon
(Table~\ref{tab:density}; embeddings in Fig.~\ref{fig:density_emb}). PCC
posts the best global $\rho$
(0.820) --- and simultaneously over-compresses the tightest cluster's scale by
$\approx$93$\times$ relative to the truth, crushing the dense core's internal
structure to a point. The rank-based global $\rho$ is scale-invariant, so this
destruction is invisible in the column that PCC leads; it is quantified only by
the value-based scale column and is directly visible in the Shepard density
plots (Fig.~\ref{fig:density_panels}a: PCC's near-band pairs collapse onto
the 2-D zero line while its far pairs track a tight monotone band). PCC's
near-band $\rho$ on this dataset is itself negative
($-0.028$): within the dense core, its distance ordering is anti-correlated
with the truth. toorPIA tops the composite
($\Sigma{=}8$: second-best near-band $\rho$ 0.312, second-best global $\rho$
0.815)
while holding the within-cluster scale at $\approx$0.4$\times$ (inflated but
legible --- inflation, unlike crushing, can be read back from the map); the
near-vs-global scatter (Fig.~\ref{fig:density_panels}b) shows both readings
at once. DREAMS takes second place ($\Sigma{=}6$) on the strength of the
best near-band $\rho$ (0.363) --- its PCA-regularized objective does deliver
the near-structure fidelity it advertises --- but its global $\rho$ (0.585)
places only fifth, well below toorPIA and PCC. The
neighbor-graph methods lead the reference block, as the bias analysis of
\S\ref{sec:metrics} predicts, while placing at or near the bottom of the
fixed-radius global column.

\subsection{Distinct dense clusters}
\label{sec:clusters}

\begin{table}[t]
\centering
\caption{Distinct dense clusters: distance-fidelity ranking at SNR${=}1$.\; Composite points: 1st$\to$5 \dots\ 5th$\to$1 on the full-$\rho$ order and on the near-$\rho$ order; rows sorted by $\Sigma$. $\rho$ columns are Shepard (Spearman) correlations vs.-ambient. \textbf{Bold} = best in column, \textit{italic} = worst; \textsuperscript{\dag} = outright-failure flag (negative near-band $\rho$, or worst tight-cluster crush exceeding $5\times$). Brackets are bootstrap 95\% CIs over $R{=}3$ seeds; deterministic methods (PCA, Isomap, DREAMS, toorPIA) show point values. recall/trust/continuity are the variable-radius $k$-NN reference block (biased; unscored). Values transcribed verbatim from the v1.4.0 committed results.}
\label{tab:clusters}
\resizebox{\textwidth}{!}{%
\begin{tabular}{lccccccccc}
\toprule
method & full & near & $\Sigma$ & full $\rho$ (global) & near $\rho$ (first-mode) & scale $\times$ & recall@15 & trust@15 & cont.@15 \\
\midrule
toorPIA & \textbf{5} & 3 & \textbf{8} & \textbf{0.585} & 0.234 & 0.517 & 0.140 & 0.949 & 0.954 \\
DREAMS & 1 & \textbf{5} & 6 & 0.413 & \textbf{0.469} & 2.634 & 0.251 & \textbf{0.960} & \textbf{0.970} \\
PyMDE & 4 & \textit{0} & 4 & 0.493 {\scriptsize[0.423, 0.519]} & 0.111 {\scriptsize[0.067, 0.118]} & 0.168 {\scriptsize[0.159, 1.378]} & \textit{0.075 {\scriptsize[0.073, 0.087]}} & \textit{0.758 {\scriptsize[0.736, 0.798]}} & \textit{0.827 {\scriptsize[0.782, 0.864]}} \\
PCC & 3 & 1 & 4 & 0.476 {\scriptsize[0.456, 0.480]} & 0.132 {\scriptsize[0.132, 0.171]} & 8.857 {\scriptsize[6.598, 10.135]}\,\textsuperscript{\dag} & 0.123 {\scriptsize[0.123, 0.128]} & 0.945 & 0.947 {\scriptsize[0.947, 0.948]} \\
PCA & 2 & 2 & 4 & 0.464 & 0.135 & \textbf{0.882} & 0.127 & 0.915 & 0.938 \\
t-SNE & \textit{0} & 4 & 4 & 0.367 {\scriptsize[0.364, 0.374]} & 0.393 {\scriptsize[0.391, 0.394]} & 1.600 {\scriptsize[1.587, 1.643]} & \textbf{0.270 {\scriptsize[0.263, 0.271]}} & 0.959 & 0.967 {\scriptsize[0.967, 0.968]} \\
UMAP & \textit{0} & \textit{0} & \textit{0} & 0.359 {\scriptsize[0.345, 0.366]} & 0.093 {\scriptsize[0.087, 0.103]} & 3.281 {\scriptsize[3.124, 3.512]} & 0.196 {\scriptsize[0.189, 0.198]} & 0.949 {\scriptsize[0.948, 0.949]} & 0.956 {\scriptsize[0.956, 0.957]} \\
Isomap & \textit{0} & \textit{0} & \textit{0} & \textit{0.337} & -0.056\,\textsuperscript{\dag} & 0.440 & 0.089 & 0.869 & 0.885 \\
\bottomrule
\end{tabular}}%
\end{table}

\begin{figure}[t]
  \centering
  \begin{subfigure}{0.58\linewidth}
    \includegraphics[width=\linewidth]{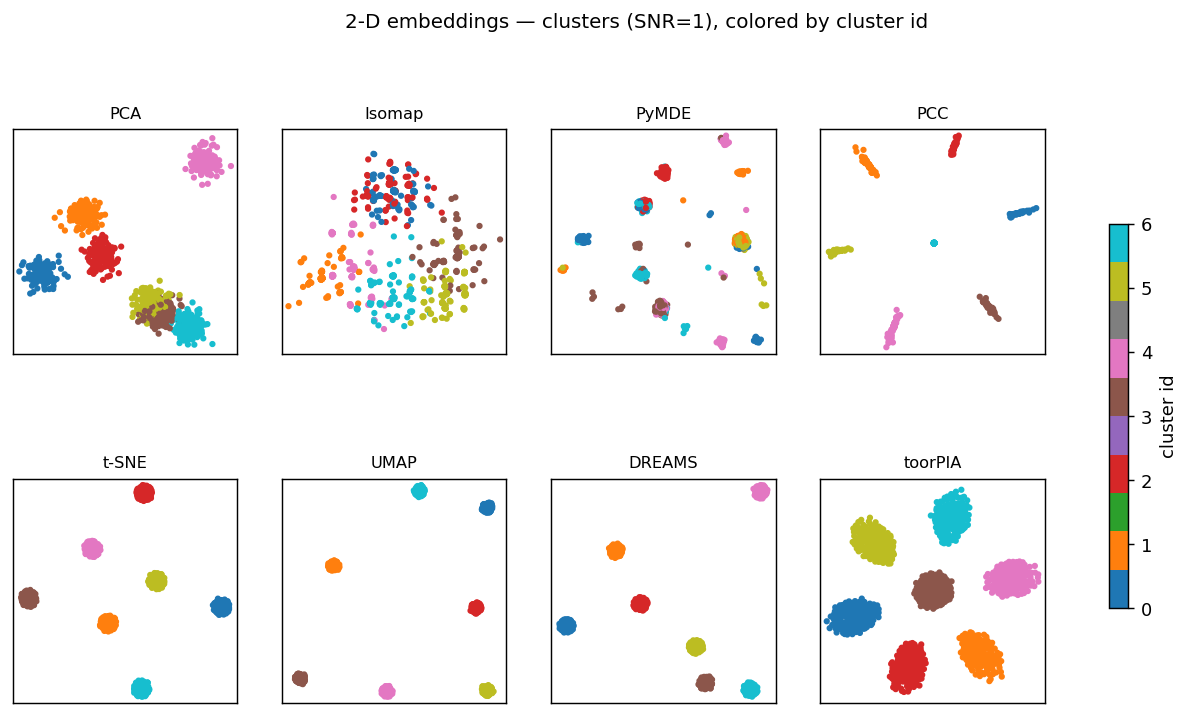}
    \caption{Embeddings, colored by cluster id.}
  \end{subfigure}\hfill
  \begin{subfigure}{0.40\linewidth}
    \includegraphics[width=\linewidth]{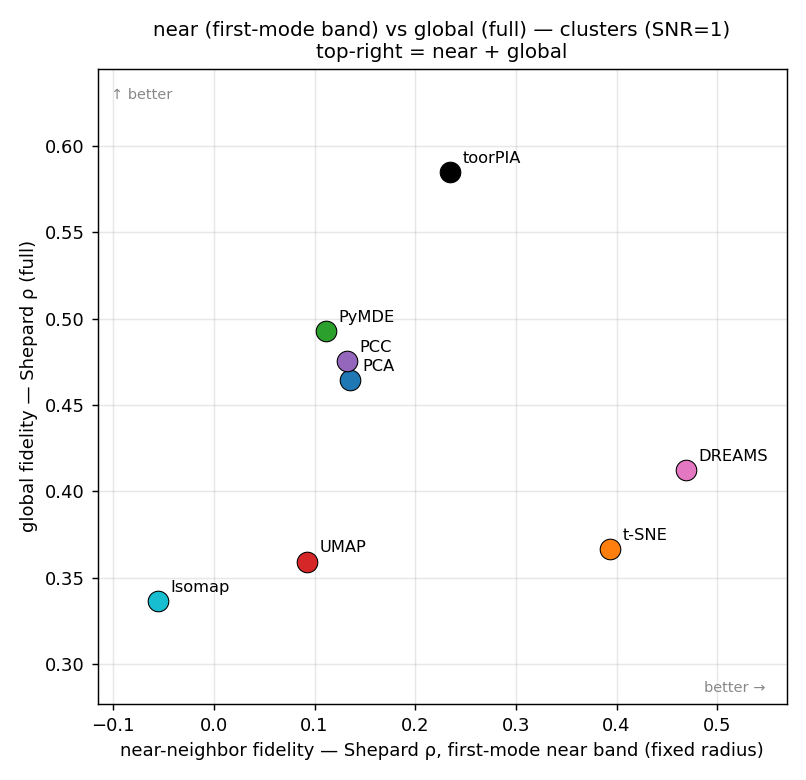}
    \caption{Near-band $\rho$ vs.\ global $\rho$.}
  \end{subfigure}
  \caption{Clusters dataset (SNR${=}1$).}
  \label{fig:clusters_panels}
\end{figure}

Seven small dense Gaussian clusters on mutually orthogonal axes (the global
geometry spans six affine dimensions, so a 2-D linear projection cannot win by
construction). toorPIA tops the composite ($\Sigma{=}8$: best global $\rho$
0.585, third-best near band); DREAMS leads the near band (0.469, ahead of
t-SNE's 0.393) and places second ($\Sigma{=}6$) despite a fifth-place global
$\rho$ (0.413); PCC again pairs a competitive global $\rho$ with an
$\approx$9$\times$ crush of the tightest cluster (Table~\ref{tab:clusters}).
Isomap's near-band $\rho$ is negative ($-0.056$) --- an outright failure
flag: within-cluster distance ordering is anti-correlated with the truth even
though its global layout is mid-pack. The gallery and the near-vs-global
scatter (Fig.~\ref{fig:clusters_panels}) show the same trade-offs
qualitatively: toorPIA draws seven separated clusters with visible internal
extent, t-SNE, UMAP, and DREAMS draw them as compact dots, PCC as needle-thin
streaks (the crush), and Isomap merges them into one cloud.

\subsection{Continuous closed-loop transition}
\label{sec:transition}

\begin{table}[t]
\centering
\caption{Continuous closed-loop transition: distance-fidelity ranking at SNR${=}1$.\; Composite points: 1st$\to$5 \dots\ 5th$\to$1 on the full-$\rho$ order and on the near-$\rho$ order; rows sorted by $\Sigma$. $\rho$ columns are Shepard (Spearman) correlations vs.-ambient. \textbf{Bold} = best in column, \textit{italic} = worst; \textsuperscript{\dag} = outright-failure flag (negative near-band $\rho$, or worst tight-cluster crush exceeding $5\times$). Brackets are bootstrap 95\% CIs over $R{=}3$ seeds; deterministic methods (PCA, Isomap, DREAMS, toorPIA) show point values. recall/trust/continuity are the variable-radius $k$-NN reference block (biased; unscored). Values transcribed verbatim from the v1.4.0 committed results.}
\label{tab:transition}
\resizebox{\textwidth}{!}{%
\begin{tabular}{lccccccccc}
\toprule
method & full & near & $\Sigma$ & full $\rho$ (global) & near $\rho$ (first-mode) & scale $\times$ & recall@15 & trust@15 & cont.@15 \\
\midrule
PCC & 4 & 3 & \textbf{7} & 0.640 {\scriptsize[0.639, 0.648]} & 0.727 {\scriptsize[0.725, 0.729]} & 2.112 {\scriptsize[1.791, 2.140]} & 0.204 {\scriptsize[0.203, 0.209]} & 0.947 {\scriptsize[0.947, 0.948]} & 0.962 {\scriptsize[0.962, 0.964]} \\
t-SNE & 2 & 4 & 6 & 0.556 {\scriptsize[0.501, 0.559]} & 0.753 {\scriptsize[0.747, 0.754]} & \textbf{1.022 {\scriptsize[0.962, 1.042]}} & \textbf{0.363 {\scriptsize[0.363, 0.364]}} & \textbf{0.974} & 0.976 {\scriptsize[0.974, 0.976]} \\
toorPIA & \textbf{5} & \textit{0} & 5 & \textbf{0.729} & 0.547 & 0.394 & 0.234 & 0.964 & 0.965 \\
DREAMS & \textit{0} & \textbf{5} & 5 & 0.537 & \textbf{0.769} & 1.136 & 0.357 & 0.974 & \textbf{0.977} \\
PCA & 3 & 1 & 4 & 0.556 & 0.630 & 0.859 & 0.182 & 0.948 & 0.956 \\
UMAP & \textit{0} & 2 & 2 & \textit{0.483 {\scriptsize[0.473, 0.491]}} & 0.695 {\scriptsize[0.694, 0.699]} & 2.410 {\scriptsize[2.323, 2.490]} & 0.292 {\scriptsize[0.288, 0.296]} & 0.966 {\scriptsize[0.966, 0.967]} & 0.968 {\scriptsize[0.968, 0.968]} \\
PyMDE & 1 & \textit{0} & 1 & 0.540 {\scriptsize[0.423, 0.581]} & \textit{0.310 {\scriptsize[0.130, 0.446]}} & 0.128 {\scriptsize[0.104, 1.777]} & \textit{0.155 {\scriptsize[0.142, 0.157]}} & \textit{0.823 {\scriptsize[0.810, 0.833]}} & \textit{0.841 {\scriptsize[0.711, 0.853]}} \\
Isomap & \textit{0} & \textit{0} & \textit{0} & 0.533 & 0.613 & 0.976 & 0.204 & 0.960 & 0.961 \\
\bottomrule
\end{tabular}}%
\end{table}

\begin{figure}[t]
  \centering
  \includegraphics[width=\linewidth]{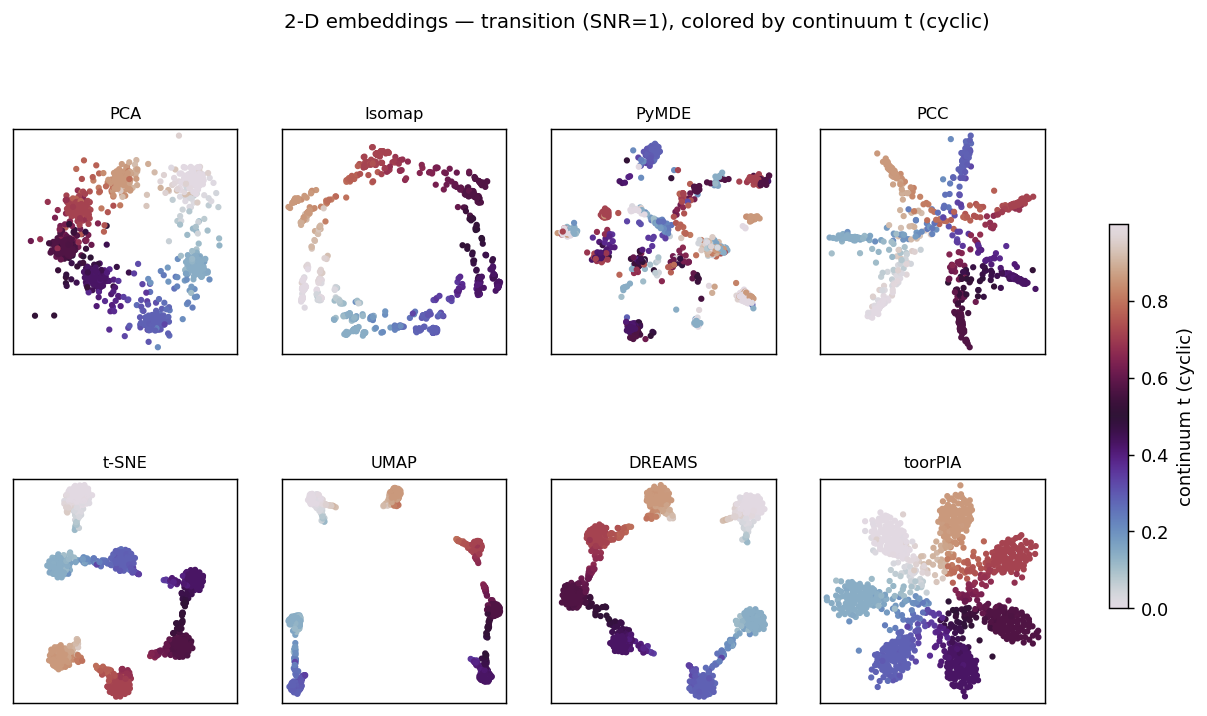}
  \caption{Transition dataset embeddings (SNR${=}1$), colored by the cyclic
  continuum parameter $t$. The defining geometry has two simultaneous
  features: seven dense typical-state clusters and their closed ring
  connectivity ($0{\to}1{\to}\dots{\to}6{\to}0$). toorPIA alone reproduces
  both; DREAMS comes closest among the rest --- dense states in the correct
  cyclic arrangement, but with torn bridges (3 of 7; see the bottleneck-gap
  diagnostic in \S\ref{sec:transition}).}
  \label{fig:transition_emb}
\end{figure}

Seven dense typical-state clusters connected into a closed loop by
heterogeneous transition bridges. This is the dataset on which toorPIA does
\emph{not} win the composite: PCC is first ($\Sigma{=}7$) and t-SNE second
($\Sigma{=}6$), while toorPIA holds the best global $\rho$ (0.729) but its
first-mode near $\rho$ ranks seventh, dropping it to joint third with DREAMS
($\Sigma{=}5$; DREAMS holds the best near $\rho$, 0.769, with a mid-pack
global 0.537) (Table~\ref{tab:transition}). The embedding gallery
(Fig.~\ref{fig:transition_emb}) shows the qualitative geometry this trade-off
hides: the dataset's defining structure has two features at once --- seven
dense states and their closed ring connectivity --- and toorPIA is the only
method that renders both simultaneously. Isomap draws the cleanest closed ring
but smears the dense clusters along it; PCA keeps the cyclic order in a ragged
ring with blurred clusters; t-SNE and UMAP recover dense clusters but tear the
connecting bridges, fragmenting the loop; DREAMS recovers the dense states
\emph{and}, through its PCA scaffold, their correct cyclic arrangement, but
it too tears bridges, leaving the ring open; PCC draws the seven states as radial
spokes fused at a central hub, so every state becomes adjacent to every other
and the cyclic adjacency is lost.

The connectivity reading is made quantitative by a committed
\emph{bridge bottleneck-gap} diagnostic (\texttt{run/bridge\_gaps.py},
\texttt{results/bridge\_gaps.csv}): for each cyclically adjacent state pair,
the smallest radius at which the two clusters become single-linkage-connected
through that segment's own bridge points (the minimax edge of the pooled
minimum spanning tree), reported as a fraction of the pair's inter-cluster
distance. Among the four methods that draw seven dense, \emph{separated}
clusters, toorPIA keeps every bridge connected --- its largest void is 14\%
of the inter-cluster distance --- while t-SNE tears 4 of the 7 bridges (voids
up to 49\%), UMAP 4 (up to 85\%), and DREAMS 3 (up to 58\%). The low gaps of
PCA/Isomap/PyMDE/PCC are connectivity-by-blurring (overlapping, smeared, or
hub-fused clusters), which the tables and the gallery expose; connectivity is
necessary for the ring reading, not sufficient. We report both readings ---
the quantitative
composite in which toorPIA is joint third, and the qualitative gallery plus
the gap diagnostic in which it is
the only method preserving the loop --- and note that the bridges (plus the
SNR${=}1$ noise) dilute the over-compression effect here: PCC's clusters are
squeezed ($\approx$2$\times$) but not to points.

\subsection{Off-subspace outliers (single-point separation)}
\label{sec:outliers}

\begin{table}[t]
\centering
\caption{Off-subspace outliers: distance-fidelity ranking at SNR${=}1$.\; Composite points: 1st$\to$5 \dots\ 5th$\to$1 on the full-$\rho$ order and on the near-$\rho$ order and on the outlier-$\rho$ order; rows sorted by $\Sigma$. $\rho$ columns are Shepard (Spearman) correlations vs.-ambient. \textbf{Bold} = best in column, \textit{italic} = worst; \textsuperscript{\dag} = outright-failure flag (negative near-band $\rho$, or worst tight-cluster crush exceeding $5\times$). Brackets are bootstrap 95\% CIs over $R{=}3$ seeds; deterministic methods (PCA, Isomap, DREAMS, toorPIA) show point values. recall/trust/continuity are the variable-radius $k$-NN reference block (biased; unscored). Values transcribed verbatim from the v1.4.0 committed results.}
\label{tab:outliers}
\resizebox{\textwidth}{!}{%
\begin{tabular}{lccccccccccc}
\toprule
method & full & near & outl. & $\Sigma$ & full $\rho$ (global) & near $\rho$ (first-mode) & scale $\times$ & recall@15 & trust@15 & cont.@15 & outlier $\rho$ \\
\midrule
toorPIA & \textbf{5} & 3 & \textbf{5} & \textbf{13} & \textbf{0.768} & 0.257 & \textbf{0.756} & 0.127 & 0.927 & 0.933 & \textbf{0.649} \\
PCA & 3 & 2 & 4 & 9 & 0.585 & 0.249 & 1.482 & 0.104 & 0.845 & 0.906 & 0.156 \\
PCC & 4 & 1 & 3 & 8 & 0.707 {\scriptsize[0.696, 0.708]} & 0.248 {\scriptsize[0.242, 0.248]} & 2.089 {\scriptsize[2.025, 2.171]} & 0.125 {\scriptsize[0.125, 0.127]} & 0.926 {\scriptsize[0.926, 0.926]} & 0.933 & 0.154 {\scriptsize[0.114, 0.164]} \\
DREAMS & 1 & \textbf{5} & \textit{0} & 6 & 0.515 & \textbf{0.477} & 3.978 & 0.223 & 0.944 & \textbf{0.956} & 0.055 \\
t-SNE & \textit{0} & 4 & 1 & 5 & 0.481 {\scriptsize[0.480, 0.502]} & 0.403 {\scriptsize[0.401, 0.406]} & 2.417 {\scriptsize[2.397, 2.434]} & \textbf{0.247 {\scriptsize[0.246, 0.250]}} & \textbf{0.944 {\scriptsize[0.943, 0.945]}} & 0.952 {\scriptsize[0.951, 0.952]} & 0.083 {\scriptsize[0.047, 0.136]} \\
PyMDE & 2 & \textit{0} & \textit{0} & 2 & 0.527 {\scriptsize[0.506, 0.580]} & 0.134 {\scriptsize[0.115, 0.141]} & 0.722 {\scriptsize[0.359, 2.135]} & 0.095 {\scriptsize[0.090, 0.096]} & \textit{0.837 {\scriptsize[0.823, 0.846]}} & \textit{0.797 {\scriptsize[0.777, 0.829]}} & \textit{0.033 {\scriptsize[0.006, 0.035]}} \\
Isomap & \textit{0} & \textit{0} & 2 & 2 & 0.490 & \textit{0.016} & 1.887 & \textit{0.089} & 0.907 & 0.912 & 0.153 \\
UMAP & \textit{0} & \textit{0} & \textit{0} & \textit{0} & \textit{0.464 {\scriptsize[0.459, 0.475]}} & 0.158 {\scriptsize[0.156, 0.166]} & 4.894 {\scriptsize[4.578, 4.964]} & 0.182 {\scriptsize[0.180, 0.182]} & 0.929 {\scriptsize[0.929, 0.930]} & 0.941 {\scriptsize[0.940, 0.941]} & 0.062 {\scriptsize[-0.005, 0.111]} \\
\bottomrule
\end{tabular}}%
\end{table}

A bulk of five dense clusters plus three anomalous \emph{directions} $\times$
two near-duplicate outliers each, placed at $3\,R_g$ along dedicated latent
axes orthogonal to the entire subspace the bulk spans --- the geometry of a
sample acquired under a different condition. The single-point question
(does one far-away point keep its separation margin?) is scored by the same
standard statistic used everywhere else, restricted by \emph{endpoint
membership} instead of by distance percentile: the anomaly-pair Shepard $\rho$
(``outlier $\rho$''), which feeds the composite as a third $5..1$ column so
that no local reading can outrank it (Table~\ref{tab:outliers}). toorPIA is
clearly first (outlier $\rho$ 0.649, deterministic --- $\approx$4$\times$ the
runner-up PCA at 0.156), pairs each same-kind anomaly co-directionally
(pair angle $\le$10$^\circ$), and holds the best global $\rho$ (0.768). The
neighbor-graph methods sit at the bottom (t-SNE 0.083, UMAP 0.062): their
anomalies land at or inside bulk clusters --- t-SNE additionally fuses each
same-kind pair into a single point, a failure that a local $k$-NN reading
scores as success. DREAMS again takes the near band (0.477) --- and its
outlier $\rho$ of 0.055 is next to last: the PCA regularization does not
rescue the single-point question, because a lone off-subspace point carries
negligible weight in both of DREAMS's objective terms, and its anomalies are
drawn amid the bulk clusters. PCC keeps the bulk clean but tears same-kind
pairs apart
(angles $\approx$86$^\circ$); PCA and Isomap drop the anomalies onto the bulk.
Few pairs involve an anomaly, so the all-pair global $\rho$ barely moves under
these failures --- which is exactly why the restricted $\rho$ is scored.

\begin{figure}[t]
  \centering
  \begin{subfigure}{0.58\linewidth}
    \includegraphics[width=\linewidth]{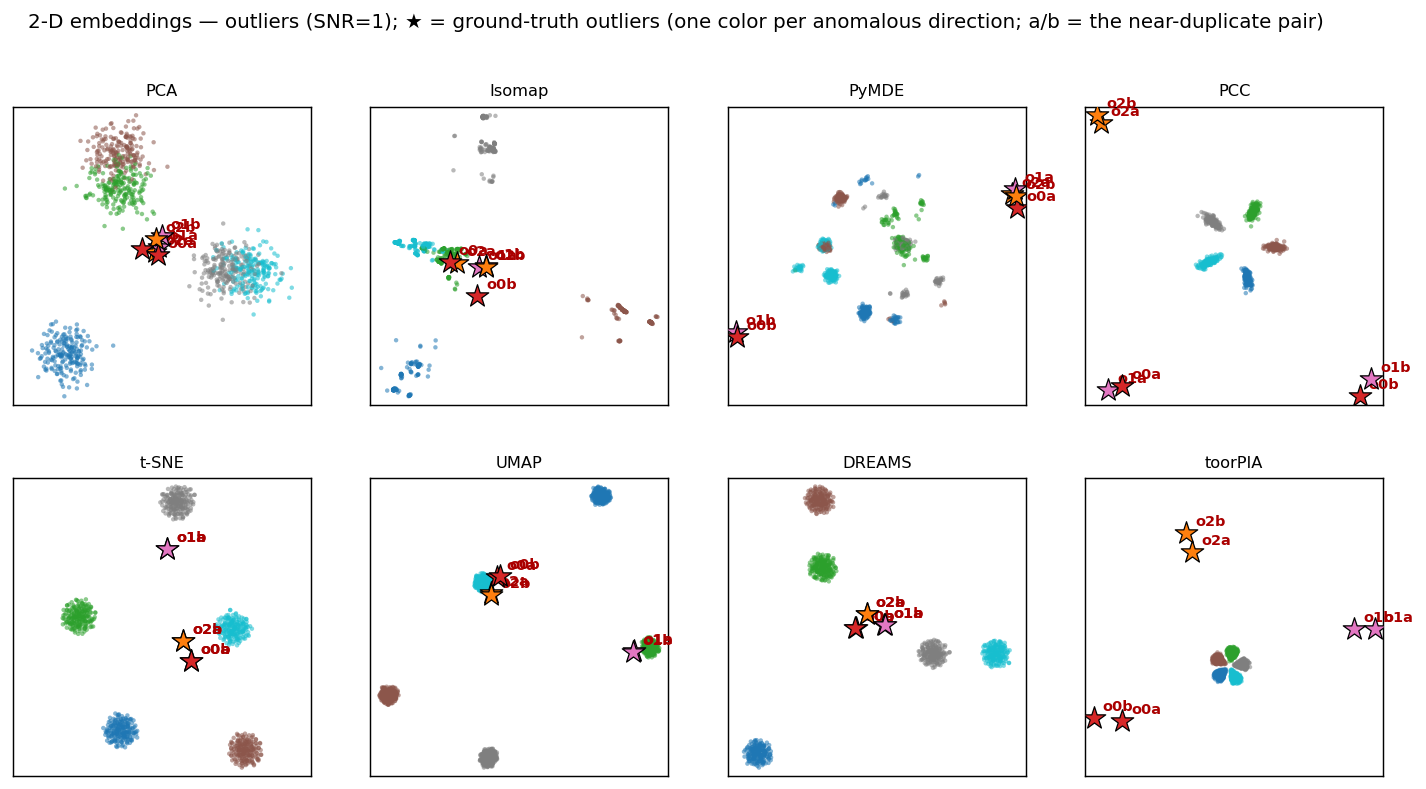}
    \caption{Embedding gallery; ground-truth outliers starred, one color per
    anomalous direction, a/b $=$ the near-duplicate pair.}
  \end{subfigure}\hfill
  \begin{subfigure}{0.40\linewidth}
    \includegraphics[width=\linewidth]{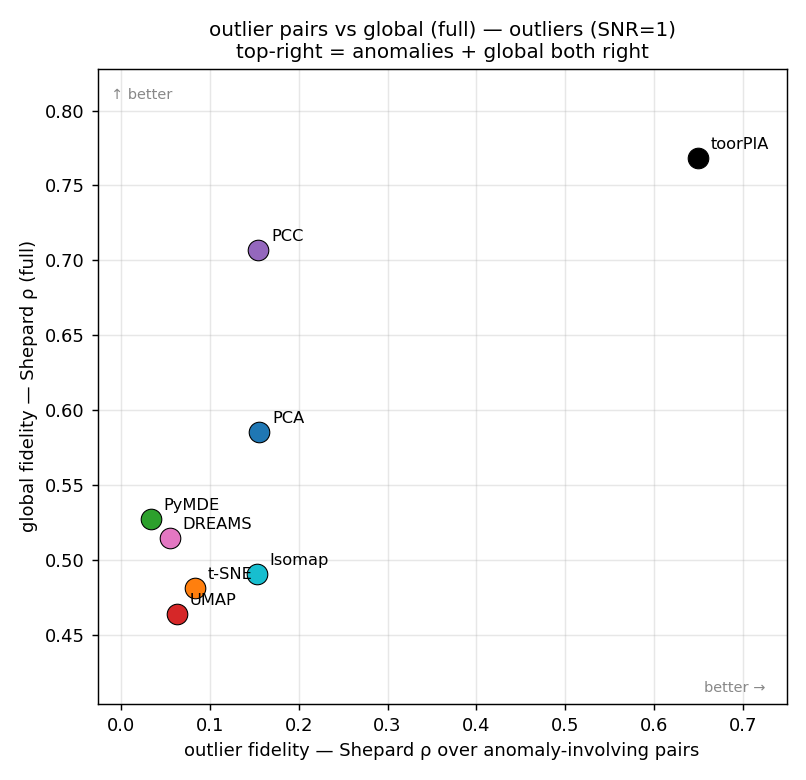}
    \caption{Anomaly-pair $\rho$ vs.\ global $\rho$; top-right renders both
    the anomalies and the global layout correctly.}
  \end{subfigure}
  \caption{Outliers dataset (SNR${=}1$, outlier factor 3).}
  \label{fig:outliers}
\end{figure}

\subsection{Imbalanced two populations (minority-structure preservation)}
\label{sec:populations}

\begin{table}[t]
\centering
\caption{Imbalanced two populations (95\% vs.\ 5\%): distance-fidelity ranking at SNR${=}1$.\; Composite points: 1st$\to$5 \dots\ 5th$\to$1 on the full-$\rho$ order and on the near-$\rho$ order and on the minority-$\rho$ order; rows sorted by $\Sigma$. $\rho$ columns are Shepard (Spearman) correlations vs.-ambient. \textbf{Bold} = best in column, \textit{italic} = worst; \textsuperscript{\dag} = outright-failure flag (negative near-band $\rho$, or worst tight-cluster crush exceeding $5\times$). Brackets are bootstrap 95\% CIs over $R{=}3$ seeds; deterministic methods (PCA, Isomap, DREAMS, toorPIA) show point values. recall/trust/continuity are the variable-radius $k$-NN reference block (biased; unscored). Values transcribed verbatim from the v1.4.0 committed results.}
\label{tab:populations}
\resizebox{\textwidth}{!}{%
\begin{tabular}{lccccccccccc}
\toprule
method & full & near & minor. & $\Sigma$ & full $\rho$ (global) & near $\rho$ (first-mode) & scale $\times$ & recall@15 & trust@15 & cont.@15 & minority $\rho$ \\
\midrule
toorPIA & \textbf{5} & 3 & 4 & \textbf{12} & \textbf{0.821} & 0.251 & 0.726 & 0.143 & 0.935 & 0.937 & 0.685 \\
DREAMS & 3 & \textbf{5} & 2 & 10 & 0.638 & \textbf{0.469} & 3.322 & 0.253 & \textbf{0.949} & \textbf{0.961} & 0.137 \\
PCC & 4 & 2 & 3 & 9 & 0.761 {\scriptsize[0.753, 0.764]} & 0.225 {\scriptsize[0.222, 0.234]} & 2.496 {\scriptsize[2.309, 2.650]} & 0.133 {\scriptsize[0.132, 0.134]} & 0.933 {\scriptsize[0.930, 0.935]} & 0.923 {\scriptsize[0.922, 0.936]} & 0.211 {\scriptsize[0.205, 0.255]} \\
PCA & 2 & 1 & \textbf{5} & 8 & 0.608 & 0.136 & 1.273 & \textit{0.080} & \textit{0.771} & 0.867 & \textbf{0.706} \\
t-SNE & \textit{0} & 4 & \textit{0} & 4 & 0.507 {\scriptsize[0.484, 0.518]} & 0.371 {\scriptsize[0.365, 0.375]} & 2.415 {\scriptsize[2.355, 2.438]} & \textbf{0.275 {\scriptsize[0.275, 0.276]}} & 0.948 & 0.957 {\scriptsize[0.956, 0.957]} & 0.070 {\scriptsize[0.047, 0.110]} \\
PyMDE & 1 & \textit{0} & 1 & 2 & 0.595 {\scriptsize[0.594, 0.666]} & 0.103 {\scriptsize[0.089, 0.125]} & 0.343 {\scriptsize[0.299, 3.383]} & 0.104 {\scriptsize[0.103, 0.105]} & 0.853 {\scriptsize[0.817, 0.869]} & \textit{0.773 {\scriptsize[0.756, 0.817]}} & 0.087 {\scriptsize[0.041, 0.124]} \\
UMAP & \textit{0} & \textit{0} & \textit{0} & \textit{0} & 0.445 {\scriptsize[0.411, 0.466]} & 0.108 {\scriptsize[0.104, 0.115]} & 5.407 {\scriptsize[4.442, 5.771]}\,\textsuperscript{\dag} & 0.200 {\scriptsize[0.199, 0.202]} & 0.934 {\scriptsize[0.931, 0.934]} & 0.933 {\scriptsize[0.933, 0.935]} & \textit{0.046 {\scriptsize[-0.033, 0.055]}} \\
Isomap & \textit{0} & \textit{0} & \textit{0} & \textit{0} & \textit{0.346} & -0.022\,\textsuperscript{\dag} & \textbf{0.913} & 0.096 & 0.883 & 0.879 & 0.079 \\
\bottomrule
\end{tabular}}%
\end{table}

A majority population (five dense clusters) and a 5\% minority with the same
internal five-cluster geometry, in disjoint latent blocks, every
cross-population center distance exactly twice the within-population one ---
the ubiquitous situation of a dominant population mixed with a small second
population whose composition is unknown in advance. Extracting the minority
from the map requires two readings positive \emph{at once}: the minority drawn
as a recognizable separate group (cross-population $\rho$) and a trustworthy
internal structure (minority-internal $\rho$). At 5\%, only toorPIA holds both
clearly positive (minority-internal 0.244, cross-population 0.660, best global
$\rho$ 0.821, deterministic). PCA places the minority correctly
(cross-population 0.683, the best) but its internal structure is gone
(minority-internal $-0.025$: a correctly placed featureless blob). t-SNE and
UMAP are the mirror image --- strong minority internals (0.658 / 0.451) with
cross-population $\rho \approx 0$: the minority's placement carries no
distance information. DREAMS inherits exactly t-SNE's failure mode at a
higher score: minority-internal 0.652 with cross-population 0.068 --- the
minority is drawn as a coherent group flung to an arbitrary position --- and
because its near-band $\rho$ (0.469) again leads, DREAMS \emph{ties} toorPIA
on the near + global columns ($5{+}3$ vs.\ $3{+}5$) while failing the reading
the dataset exists to test; only the third scored column separates them
($\Sigma{=}12$ vs.\ 10). That column --- the minority-pair $\rho$ over
every pair with at least one minority endpoint --- is the single number that
drops if either reading fails (DREAMS 0.137, t-SNE 0.070, UMAP 0.046), but it
is not the full reading by itself: PCA tops it (0.706, ahead of toorPIA's
0.685) because at 5\% the minority--majority pairs make up 95\% of the
minority-involving pairs, so a correctly \emph{placed} blob with no internal
structure still scores high; the diagnostic pair, not any single composite,
separates the methods. PCC fails both (0.023 / 0.198) --- and the failure is
\emph{silent}: PCC still posts the second-best global $\rho$ (0.761), so
nothing in the global metric warns that the minority was destroyed. This is
consistent with reference-point subsampling: pairs internal to the minority
carry a share of the loss that shrinks with the \emph{square} of the minority
fraction (0.25\% of terms at 5\%), so minority points are placed almost
entirely by their relations to majority reference points.

\begin{figure}[t]
  \centering
  \begin{subfigure}{0.58\linewidth}
    \includegraphics[width=\linewidth]{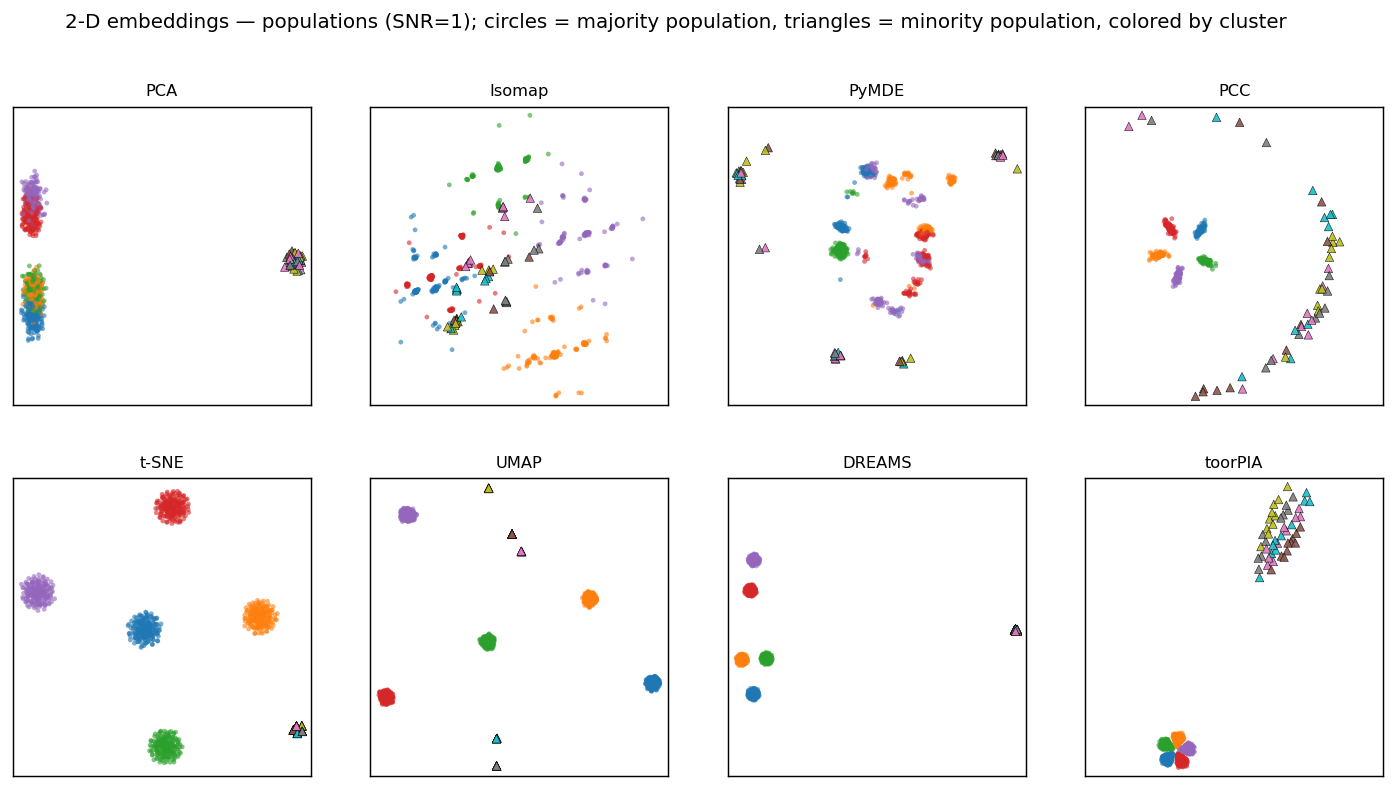}
    \caption{Gallery: circles $=$ majority, triangles $=$ minority, colored by
    cluster.}
  \end{subfigure}\hfill
  \begin{subfigure}{0.40\linewidth}
    \includegraphics[width=\linewidth]{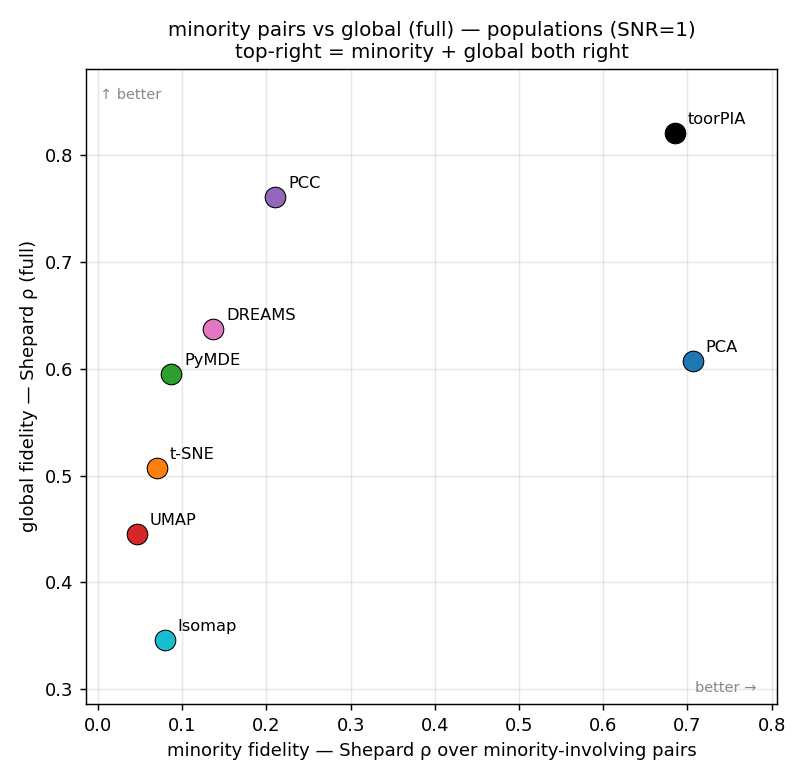}
    \caption{Minority-pair $\rho$ vs.\ global $\rho$.}
  \end{subfigure}
  \caption{Imbalanced populations dataset at 95\% vs.\ 5\% (SNR${=}1$).}
  \label{fig:populations}
\end{figure}

\subsection{Run-to-run stability (Procrustes)}
\label{sec:stability}
Stochastic methods are run with $R{=}3$ seeds; run-to-run wobble is
summarized as the per-point positional dispersion after removing the
rotation/scale/flip gauge (Procrustes) together with the across-seed spread of
the headline metrics, and is committed per dataset in the artifact's
\texttt{results/stability.csv}. PCA, Isomap, DREAMS, and toorPIA are
deterministic
(same input $\to$ identical map; toorPIA's embedding endpoint exposes no
seed, and DREAMS's fixed PCA initialization leaves its single-threaded
optimizer no randomness). The stochastic methods' wobble is small relative to the between-method
differences discussed above, and the bootstrap CIs shown in every ranking
table carry the seed-level uncertainty into the comparisons; we assert no
strict winner where CIs overlap.

\section{Supplement: out-of-sample monitoring (addplot)}
\label{sec:addplot}

\begin{table}[t]
\centering
\caption{Out-of-sample (addplot) monitoring test at SNR${=}1$: cluster-anchored anomalies (3\,Rg along dimensions the normal basemap never varies in) and 50 fresh normal controls, added one at a time to a basemap fitted on normal data only. Detection $=$ anomaly distance from the map centroid over the bulk's median radius (large $=$ visibly outside); attribution $=$ direction from the centroid identifies the source cluster. PCA/Isomap use \texttt{transform}; UMAP a seeded \texttt{transform}; DREAMS openTSNE's partial-optimization \texttt{transform} (one point per call; its regularization acts only at fit time); toorPIA server-side \texttt{addplot\_embedding}. t-SNE, PyMDE, and PCC expose no out-of-sample operation. Values transcribed verbatim from the v1.4.0 committed results.}
\label{tab:addplot}
\resizebox{\textwidth}{!}{%
\begin{tabular}{lcccccc}
\toprule
method & anom.\ dist $\div$ bulk radius (med) & min & attribution acc. & angle to own cluster $^\circ$ (med) & pair angle $^\circ$ (med) & control ratio ($\approx$1 ideal) \\
\midrule
PCA & 0.966 & 0.687 & 0.800 & 3.310 & 5.025 & 0.874 \\
Isomap & 1.337 & 0.755 & 0.900 & 7.203 & 1.932 & 1.011 \\
PyMDE & \multicolumn{6}{l}{\textit{not operable: pymde optimizes fit coordinates; no transform}} \\
PCC & \multicolumn{6}{l}{\textit{not operable: pccdr optimizes fit coordinates; no transform}} \\
t-SNE & \multicolumn{6}{l}{\textit{not operable: sklearn TSNE has no out-of-sample transform}} \\
UMAP & 0.971 {\scriptsize[0.962, 1.042]} & 0.281 {\scriptsize[0.261, 0.486]} & 1.000 & 3.334 {\scriptsize[2.176, 3.429]} & 1.559 {\scriptsize[1.220, 3.801]} & 0.939 {\scriptsize[0.935, 1.003]} \\
DREAMS & 1.037 & 0.533 & 1.000 & 0.530 & 0.293 & 1.070 \\
toorPIA & 9.732 & 8.654 & 1.000 & 0.890 & 0.630 & 0.959 \\
\bottomrule
\end{tabular}}%
\end{table}

The main benchmark embeds a fixed dataset; industrial monitoring poses a
different, operational question. The basemap is fitted on \emph{normal} data
only, and new points arrive afterwards, one at a time. The added set holds
cluster-anchored anomalies --- each shares a normal cluster's profile in the
measured features and deviates $3\,R_g$ along new dimensions orthogonal to
everything the normal data varies in (a near-duplicate pair per cluster,
5 clusters $\times$ 2) --- the realistic shape of a fault: a known operating
state plus an effect the historical data never showed --- plus 50 fresh normal
points as controls. Two questions, in order: \emph{detection} --- does the
anomaly land visibly outside the normal region at all? --- and
\emph{attribution} --- does its direction from the map centroid identify the
source cluster? The ambient high-D features resolve attribution 10/10 (the
anchor signal survives the SNR${=}1$ noise), so a faithful map can too.

Three of the eight methods --- t-SNE (as implemented in scikit-learn), PyMDE,
and PCC --- expose no out-of-sample operation at all; for monitoring, that is
itself the finding: adding data means re-fitting, and a re-fit re-arranges the
map. Among the operable methods (Table~\ref{tab:addplot}), only toorPIA
answers both questions: every anomaly lands far outside the normal region
(median 9.7$\times$ the bulk radius, minimum 8.7$\times$; deterministic) and
its direction identifies the source cluster (attribution 10/10, median angle
to its own cluster 0.9$^\circ$, pair angle 0.6$^\circ$). For PCA, Isomap,
UMAP, and DREAMS the anomalies land \emph{inside or at} the normal clusters
(median
radius ratios 0.97--1.34, minima down to 0.28): the anomaly is drawn as just
another normal point of its source cluster, so the nominally high attribution
accuracy (0.8--1.0) carries no alarm --- detection silently fails. DREAMS is
the sharpest instance of the pattern: the best attribution geometry of the
non-toorPIA methods (median own-cluster angle 0.5$^\circ$, pair angle
0.3$^\circ$) attached to a median radius ratio of 1.04 --- a perfect
direction with no alarm. Added
normal controls land correctly for every operable method (ratio 0.87--1.07).

Three honest notes on the protocol. (1) toorPIA's \texttt{addplot\_embedding}
targets the fitted basemap's server-side state, so this test performs one
live \texttt{basemap\_embedding} call and one \texttt{addplot\_embedding}
call per added point (the monitoring semantics: points arrive one at a time) and
commits the two coordinate sets as a self-consistent cache pair; the addplot
inherits the basemap's preprocessing server-side, so basemap and added points
receive identical treatment. (2) The PCA/Isomap/DREAMS transforms and toorPIA
are deterministic; UMAP's transform is seeded. DREAMS's out-of-sample
operation is openTSNE's partial optimization of each new point (one per call)
against the fixed basemap; its regularization term is a fit-time objective
over the full embedding and does not act at transform time. (3) A
re-fit-based alternative for
the methods without an out-of-sample operation (append the new data, re-fit,
Procrustes-align, measure displacement) is future work --- it measures a
different, weaker property (map stability under re-fit), not the monitoring
operation itself.

\begin{figure}[t]
  \centering
  \includegraphics[width=0.9\linewidth]{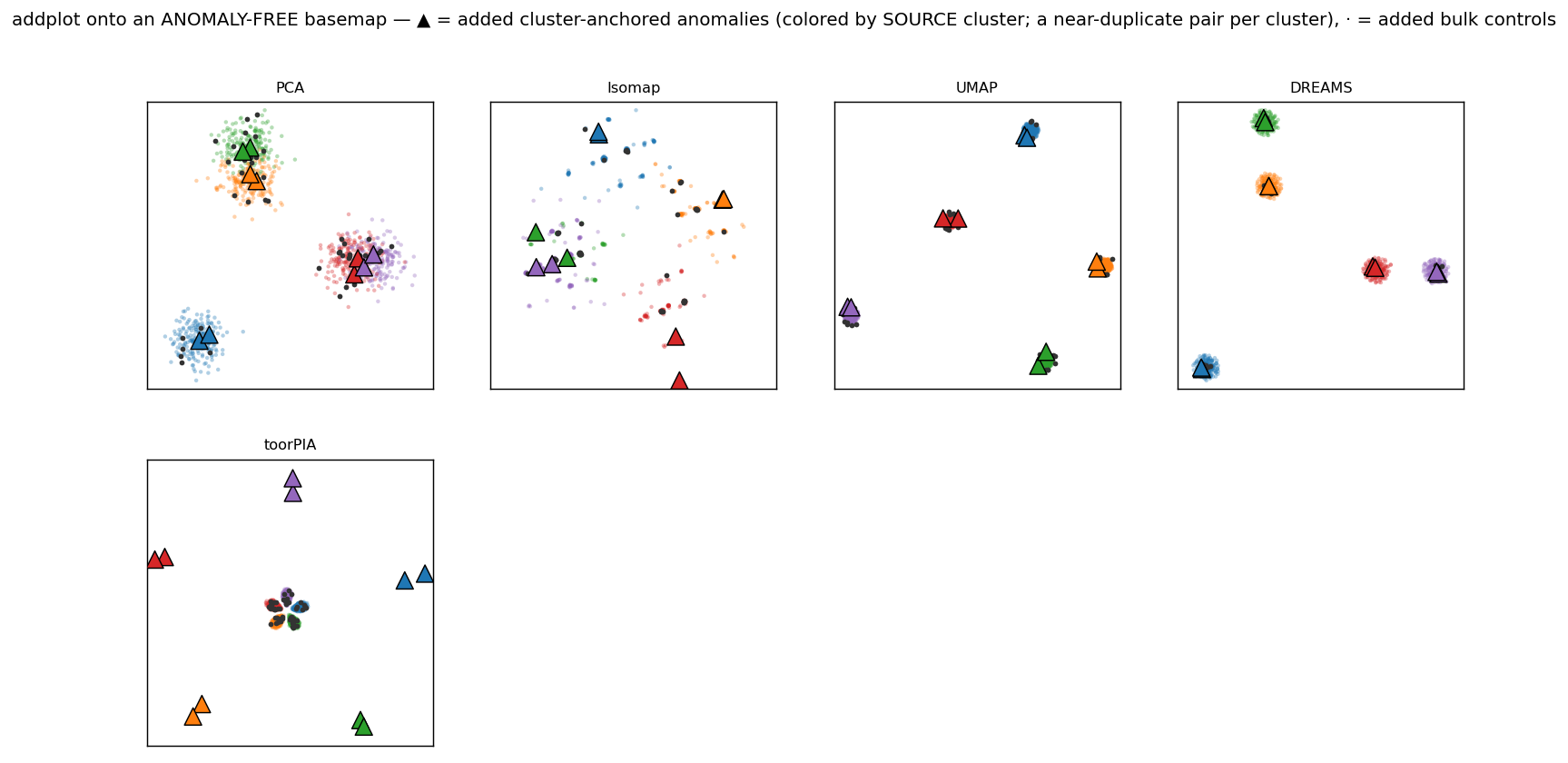}
  \caption{Anomaly-free basemap plus added points ($\blacktriangle$ =
  cluster-anchored anomalies colored by source cluster, $\cdot$ = added normal
  controls). Faithful = each $\blacktriangle$ outside the normal region and in
  its own cluster's direction; controls inside the bulk.}
  \label{fig:addplot}
\end{figure}

\section{Discussion and Limitations}
\label{sec:limitations}

\subsection{Scope}

This benchmark characterizes \emph{distance and structure preservation on
synthetic, known-structure data}. It is not a claim about any method's
superiority on real downstream tasks. That applies equally to the outliers
dataset and the addplot readouts: they characterize whether a synthetic
anomaly structure survives the 2-D map, not any method's usefulness for
real-world outlier \emph{detection}, which is a downstream task with its own
tooling. Rankings respect CI overlap throughout --- where CIs overlap, no
strict winner is asserted. All five-dataset results are obtained under the
redundancy-rich isotropic noise model, the noise-\emph{friendly} extreme
(\S\ref{sec:protocol}); the artifact's noise-dims sweep shows that rankings
need not transfer to sparse or irrelevant-feature regimes, and none of the
rankings here should be extrapolated there. Validation on real data is
future work.

\subsection{Conflict of interest and trust measures}

\textbf{Disclosure: this benchmark is maintained by the vendor of toorPIA},
and the author is affiliated with that vendor. The benchmark is therefore
designed so that its claims do not rest on the maintainer's judgment, through
four structural safeguards. (1)~\emph{Metrics are computed independently of
every method} --- exactly, on all pairwise distances, never through any
method's internal reference-point or neighbor-graph approximation (toorPIA's
included: only its output coordinates are ever seen). (2)~\emph{Hypotheses
are documented before the results}: the constraint-density hypothesis behind
the outliers dataset was committed to version control before the
corresponding result tables, and results are reported however they come out,
including when they contradict the hypothesis or are unfavorable to toorPIA.
(3)~\emph{Every number is third-party recomputable offline}
(\S\ref{sec:reproducibility}). (4)~\emph{No strict winner is asserted when
CIs overlap}, and the biased-but-standard reference metrics (the recall@$k$
family) are always reported alongside the primary ones.

The transparency this buys is visible in the results themselves: the
first-mode near band goes to DREAMS --- not toorPIA --- on every
one of the five datasets; on the transition dataset the composite goes to
PCC with toorPIA joint third (\S\ref{sec:transition}); on populations the
scored minority-pair $\rho$ goes to PCA, not toorPIA, and DREAMS ties toorPIA
on the near + global columns (\S\ref{sec:populations}); and the
hyperparameter-sensitivity
sweep discloses that tuned t-SNE overtakes the near-band $\rho$ leaders on
density and draws level with toorPIA's density composite
(\S\ref{sec:methodsconfig}). These unfavorable readings are printed
in the same tables, with the same prominence, as the favorable ones.

\subsection{Reproducibility of a closed-source method}
\label{sec:reproducibility}

toorPIA is a remote API. We never inspect its internals; we characterize its
input$\to$output behavior on data whose true structure is known, exactly as
we do for the open-source methods. Its output coordinates --- not its
algorithm --- are committed to the repository, so every toorPIA number in
this paper can be recomputed offline, by anyone, without an API key: the
metrics pipeline reads the committed coordinates and evaluates them with the
same independent machinery used for every other method. The open-source
methods are byte-reproducible end-to-end from the deterministic driver (same
arguments $\to$ identical numbers, enforced by tests), and all per-run
metric tables are committed.

The honest limitation: a third party cannot \emph{regenerate} toorPIA's
coordinates without access to the vendor's API. Independent reproduction of
toorPIA's rows is therefore limited to recomputation from the committed
cache (archived with the release tag and Zenodo DOI), plus whatever access
to the API a reader can obtain. A reader who trusts the committed cache can
verify everything; a reader who does not must treat toorPIA's rows as
vendor-supplied coordinates evaluated by open code. This asymmetry is
intrinsic to benchmarking a closed-source method and is stated here rather
than papered over.

\section{Conclusion}
\label{sec:conclusion}

We proposed scoring DR fidelity with a fixed-radius distance-band Shepard
$\rho$ --- a structure-adaptive near band (the first mode of the pairwise
distance profile) reported separately from the global number, every point
judged on the same absolute radius --- together with value-based complements
(band stress, tightest-cluster over-compression) and membership-restricted
variants for single-point and minority-population questions. The structural
argument of \S\ref{sec:metrics} explains why the field's standard local
metrics (recall@$k$, trustworthiness, continuity) are biased toward
neighbor-graph methods: a per-point variable radius plus a hard inclusion
threshold measures agreement with a $k$-NN construction, not faithful
reproduction of near distances. We keep that family as a labelled reference,
and the benchmark's results disagree with it in exactly the direction the
bias predicts.

On five known-geometry datasets at SNR${=}1$, the eight-method benchmark
shows what the composite metric set sees and single numbers hide: a
top-tier global $\rho$ coexisting with a ${\approx}93\times$ crush of the
tightest cluster's scale; a minority population destroyed while the global
$\rho$ stays second-best; anomalies drawn inside bulk clusters by exactly
the methods the $k$-NN reference favors; a recent local-plus-global hybrid
(DREAMS) that takes the near band on every dataset yet fails the same
single-point and minority-placement questions as the neighbor-graph family
--- a failure only the membership-restricted $\rho$ exposes, since on
populations its near + global columns tie the leader's; and, out of sample,
one method
alone placing a never-seen anomaly outside the normal region with a
direction that identifies its source. No method wins everything --- toorPIA
leads the composite on density, clusters, outliers, and populations, and is
joint third on
transition, while DREAMS owns the near band everywhere --- and the
benchmark reports all of it. Because every number is recomputable offline
from committed artifacts, including the closed-source method's output
coordinates, the release (v1.4.0, archived on Zenodo) is intended as a
reproducible, externally citable characterization that future work --- and
other people's papers --- can cite instead of vendor claims.

\section*{Data and Code Availability}
The benchmark code, committed results, figures, and toorPIA output-coordinate
caches are available at \url{https://github.com/toorpia/dr-fidelity-benchmark}
(release tag v1.4.0), archived at Zenodo (version DOI
\href{https://doi.org/10.5281/zenodo.21380823}{10.5281/zenodo.21380823};
all versions:
\href{https://doi.org/10.5281/zenodo.21189374}{10.5281/zenodo.21189374}).
This paper reports the results committed at that tag.

\bibliographystyle{unsrt}
\bibliography{references}

\begin{thebibliography}{10}

\bibitem{vandermaaten2008tsne}
Laurens van~der Maaten and Geoffrey Hinton.
\newblock Visualizing data using {t-SNE}.
\newblock {\em Journal of Machine Learning Research}, 9(86):2579--2605, 2008.

\bibitem{mcinnes2018umap}
Leland McInnes, John Healy, and James Melville.
\newblock {UMAP}: Uniform manifold approximation and projection for dimension reduction.
\newblock {\em arXiv preprint arXiv:1802.03426}, 2018.

\bibitem{espadoto2019survey}
Mateus Espadoto, Rafael~M. Martins, Andreas Kerren, Nina S.~T. Hirata, and Alexandru~C. Telea.
\newblock Toward a quantitative survey of dimension reduction techniques.
\newblock {\em IEEE Transactions on Visualization and Computer Graphics}, 27(3):2153--2173, 2021.

\bibitem{shepard1962}
Roger~N. Shepard.
\newblock The analysis of proximities: Multidimensional scaling with an unknown distance function. {I}.
\newblock {\em Psychometrika}, 27(2):125--140, 1962.

\bibitem{beyer1999nearest}
Kevin Beyer, Jonathan Goldstein, Raghu Ramakrishnan, and Uri Shaft.
\newblock When is ``nearest neighbor'' meaningful?
\newblock In {\em Proceedings of the 7th International Conference on Database Theory (ICDT)}, pages 217--235, 1999.

\bibitem{aggarwal2001surprising}
Charu~C. Aggarwal, Alexander Hinneburg, and Daniel~A. Keim.
\newblock On the surprising behavior of distance metrics in high dimensional space.
\newblock In {\em Proceedings of the 8th International Conference on Database Theory (ICDT)}, pages 420--434, 2001.

\bibitem{venna2006trustworthiness}
Jarkko Venna and Samuel Kaski.
\newblock Local multidimensional scaling.
\newblock {\em Neural Networks}, 19(6-7):889--899, 2006.

\bibitem{lee2009qualityassessment}
John~A. Lee and Michel Verleysen.
\newblock Quality assessment of dimensionality reduction: Rank-based criteria.
\newblock {\em Neurocomputing}, 72(7-9):1431--1443, 2009.

\bibitem{hotelling1933pca}
Harold Hotelling.
\newblock Analysis of a complex of statistical variables into principal components.
\newblock {\em Journal of Educational Psychology}, 24(6):417--441, 1933.

\bibitem{tenenbaum2000isomap}
Joshua~B. Tenenbaum, Vin de~Silva, and John~C. Langford.
\newblock A global geometric framework for nonlinear dimensionality reduction.
\newblock {\em Science}, 290(5500):2319--2323, 2000.

\bibitem{roweis2000lle}
Sam~T. Roweis and Lawrence~K. Saul.
\newblock Nonlinear dimensionality reduction by locally linear embedding.
\newblock {\em Science}, 290(5500):2323--2326, 2000.

\bibitem{belkin2003laplacian}
Mikhail Belkin and Partha Niyogi.
\newblock Laplacian eigenmaps for dimensionality reduction and data representation.
\newblock {\em Neural Computation}, 15(6):1373--1396, 2003.

\bibitem{agrawal2021pymde}
Akshay Agrawal, Alnur Ali, and Stephen Boyd.
\newblock Minimum-distortion embedding.
\newblock {\em Foundations and Trends in Machine Learning}, 14(3):211--378, 2021.

\bibitem{pcc2025}
Jacob Gildenblat and Jens Pahnke.
\newblock Preserving clusters and correlations: A dimensionality reduction method for exceptionally high global structure preservation.
\newblock {\em arXiv preprint arXiv:2503.07609}, 2025.

\bibitem{kury2026dreams}
No{\"e}l Kury, Dmitry Kobak, and Sebastian Damrich.
\newblock {DREAMS}: Preserving both local and global structure in dimensionality reduction.
\newblock {\em Transactions on Machine Learning Research}, 2026.
\newblock arXiv:2508.13747.

\bibitem{kruskal1964mds}
Joseph~B. Kruskal.
\newblock Multidimensional scaling by optimizing goodness of fit to a nonmetric hypothesis.
\newblock {\em Psychometrika}, 29(1):1--27, 1964.

\bibitem{jeon2021steadiness}
Hyeon Jeon, Hyung-Kwon Ko, Jaemin Jo, Youngtaek Kim, and Jinwook Seo.
\newblock Measuring and explaining the inter-cluster reliability of multidimensional projections.
\newblock {\em IEEE Transactions on Visualization and Computer Graphics}, 28(1):551--561, 2022.

\bibitem{bae2025metric}
Jiyeon Bae, Hyeon Jeon, and Jinwook Seo.
\newblock Metric design != metric behavior: Improving metric selection for the unbiased evaluation of dimensionality reduction.
\newblock In {\em 2025 IEEE Visualization and Visual Analytics (VIS)}, pages 46--50, 2025.

\bibitem{smelser2024stress}
Kiran Smelser, Jacob Miller, and Stephen Kobourov.
\newblock ``normalized stress'' is not normalized: How to interpret stress correctly.
\newblock In {\em 2024 IEEE Evaluation and Beyond -- Methodological Approaches for Visualization (BELIV)}, pages 41--50, 2024.

\end{thebibliography}

\end{document}